%% file: TIP2019ellipseRCNN.tex
\documentclass[journal]{IEEEtran}

\usepackage{graphicx}
\usepackage{amsmath}
\usepackage{amssymb}
\usepackage{cite}
\usepackage{array}
\usepackage{url}
\usepackage{booktabs}
\usepackage{multirow}
\usepackage{makecell}

\usepackage[dvipsnames]{xcolor}

\newcolumntype{M}[1]{>{\centering\arraybackslash}m{#1}}

\begin{document}
%
\title{Ellipse R-CNN: Learning to Infer Elliptical Object\\ from Clustering and Occlusion}
%
%
%

\author{Wenbo~Dong,~\IEEEmembership{Student~Member,~IEEE,}
        Pravakar~Roy,~\IEEEmembership{Student~Member,~IEEE,}
        Cheng~Peng,~\IEEEmembership{Student~Member,~IEEE,} and~Volkan~Isler,~\IEEEmembership{Senior~Member,~IEEE}
\thanks{W. Dong, P. Roy, C. Peng, and V. Isler are with the Department of Computer Science and Engineering, University of Minnesota, Minneapolis, MN, 55455, USA (e-mail: dongx358@umn.edu, royxx268@umn.edu, peng0175@umn.edu, isler@cs.umn.edu).}
\thanks{This work was supported in part by USDA NIFA MIN-98-G02, in part by the MnDrive initiative, and in part by NSF \#1722310.}
}

\maketitle

\begin{abstract}
Images of heavily occluded objects in cluttered scenes, such as fruit clusters in trees, are hard to segment.
To further retrieve the 3D size and 6D pose of each individual object in such cases, bounding boxes are not reliable from multiple views since only a little portion of the object's geometry is captured.
We introduce the first CNN-based ellipse detector, called Ellipse R-CNN, to represent and infer occluded objects as ellipses.
We first propose a robust and compact ellipse regression based on the Mask R-CNN architecture for elliptical object detection.
Our method can infer the parameters of multiple elliptical objects even they are occluded by other neighboring objects.
For better occlusion handling, we exploit refined feature regions for the regression stage, and integrate the U-Net structure for learning different occlusion patterns to compute the final detection score.
The correctness of ellipse regression is validated through experiments performed on synthetic data of clustered ellipses.
We further quantitatively and qualitatively demonstrate that our approach outperforms the state-of-the-art model (i.e., Mask R-CNN followed by ellipse fitting) and its three variants on both synthetic and real datasets of occluded and clustered elliptical objects.
\end{abstract}

\begin{IEEEkeywords}
Ellipse regression, occlusion handling, 3D object localization, object detection, convolutional neural networks.
\end{IEEEkeywords}

%
\IEEEpeerreviewmaketitle

\input{tipIntroduction}

\input{tipRelatedWork}

\input{tipOverviewBbox}

\input{tipEllipseRegression}

\input{tipExperiments}

\section{Conclusion}
This paper shows that traditional R-CNN methods are not well-suited for ellipse fitting since they only predict bounding boxes that have no orientation information for objects, and they are typically trained on the whole object regions in occluded cases.
This makes those deep models suffer from outputting a large number of false positives and being unreliable to serve as the inputs for further 3D estimation of the object pose and its sizes. 
We thus propose the Ellipse R-CNN to focus on the visible regions and infer the whole elliptical objects as ellipses from heavy occlusions.
A robust ellipse regression is formulated to generalize both occluded and unoccluded cases.
Our model firstly learns various occlusion patterns of ellipses within the refined visible regions, then generates the final classification score by integrating the visibility information from an attention vector and the whole object information from the regressed ellipse.
In this way, the model learns discriminative representations of occluded objects, which are robust in differently oriented scenarios.
Extensive experimental results on two synthetic datasets and two real datasets demonstrate the advantages of our model compared to the Mask R-CNN.
The current approach for 3D object estimation weights equally each predicted ellipse parameter from 2D detections.
Our future work would investigate predicting the uncertainties for all ellipse parameters to further boost the accuracy of the 3D object estimation system.


%



\section*{Acknowledgment}
We thank our colleagues Nicolai H\"{a}ni and Zhihang Deng from the University of Minnesota, for providing valuable feedback and technical support throughout this research.

\input{tipAppendices}




%
\bibliographystyle{IEEEtran}
\bibliography{tipReferences}

%
%

%

\begin{IEEEbiography}{Wenbo Dong:}
Doctor
\end{IEEEbiography}

\begin{IEEEbiographynophoto}{Pravakar Roy:}
Doctor
\end{IEEEbiographynophoto}

\begin{IEEEbiographynophoto}{Cheng Peng:}
Ph.D. candidate
\end{IEEEbiographynophoto}


\begin{IEEEbiographynophoto}{Volkan Isler:}
Professor
\end{IEEEbiographynophoto}




\end{document}

%% file: tipIntroduction.tex
\section{Introduction} \label{sec:introduction}
\begin{figure}[t]
	\centering
	\includegraphics[width=0.99\columnwidth]{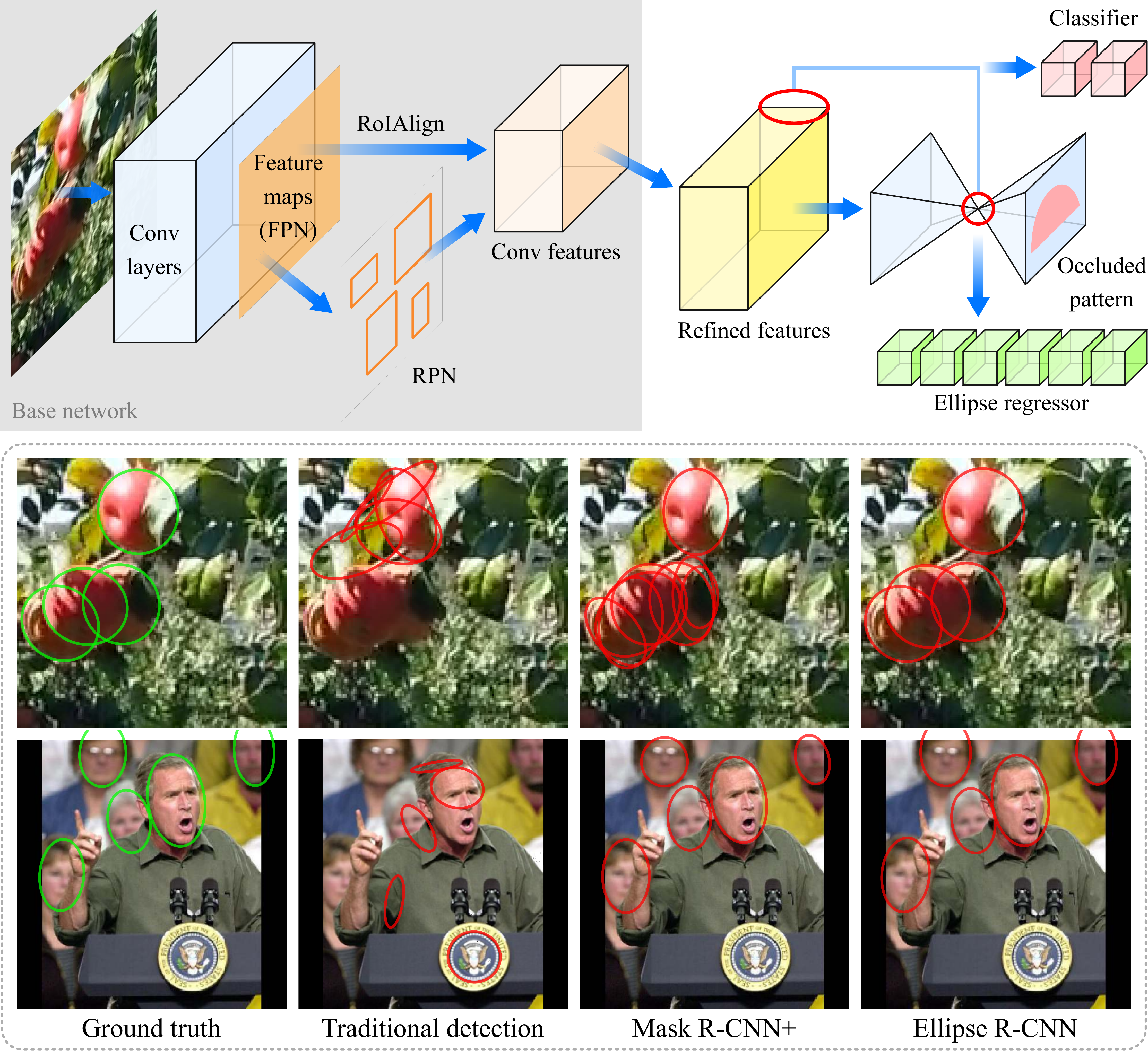}
	\caption{Overview of our proposed Ellipse R-CNN system. Lower: Sample images of occluded elliptical object detection in the clustered case. Traditional detection: traditional ellipse detection in 2D~\cite{xie2002new}. Mask R-CNN+: Directly fitting ellipses from the entire object masks output by Mask R-CNN~\cite{hemask}. Both of these two methods suffer from false positives and fail to capture the ellipse orientation. Our proposed Ellipse R-CNN outputs accurate ellipses compared to the ground truth (green colored).}
	\label{fig:motivation2D}
\end{figure}

\IEEEPARstart{D}{etection} of ellipse-like shapes~\cite{xie2002new} has been widely used in various image processing tasks, for instance, face detection~\cite{zhang2005robust} and medical imaging diagnosis~\cite{lu2008detection}.
The above works all investigate the mathematical model of ellipse based on segmented edges, contours, and curvatures~\cite{prasad2012edge} from the image to identify different ellipses.
However, such traditional methods for ellipse fitting highly rely on pre-processing (such as segmentation and grouping), and thus often fail to detect ellipsoid objects from the image, especially in complex environments (see Fig.~\ref{fig:motivation2D} and Fig.~\ref{fig:expResReal}--\ref{fig:expResFDDB}).
For fruit detection in modern orchard settings, as an example, the edge information of clustered fruits is not salient~\cite{roy2016vision} and largely interfered by nearby obstacles and background scenes, such as leaves and branches (see Fig.~\ref{fig:motivation2D}).

\begin{figure*}[!t]
	\centering
	\includegraphics[width=0.99\textwidth]{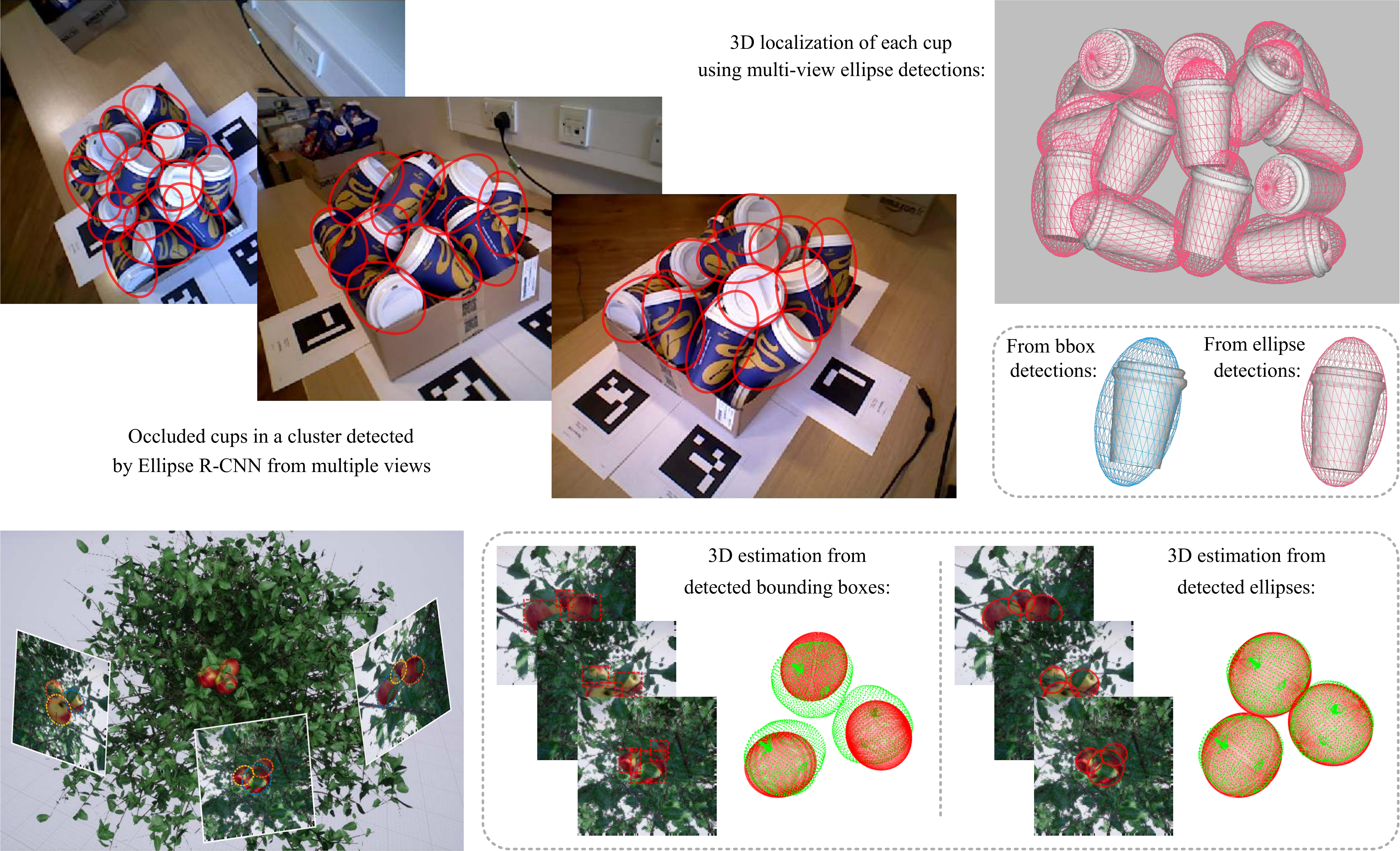}
	\caption{3D estimation of occluded elliptical objects based on multi-view image detections in semantic localization applications. Upper: Detected ellipses (red-colored) of coffee cups in a cluster~\cite{doumanoglou2016recovering} from our Ellipse R-CNN and estimated 3D ellipsoids of occluded cups. Accurate 3D estimation of ellipsoids (pink-colored) from multi-view ellipse detections better enclose the ground-truth cup models compared to bounding-box detections~\cite{nicholson2019quadricslam}. Lower: A cluster of fruit objects captured from multiple views. In the clustered and occluded case, multi-view 3D object estimation~\cite{rubino20183d} from proposed ellipse detections outputs much more accurate sizes in 3D and poses in 6D than using bounding-box constraints~\cite{nicholson2019quadricslam} (3D ground-truth points are green-colored).}
	\label{fig:motivation3D}
\end{figure*}

Adapting convolutional neural networks (CNNs) for object detection~\cite{ren2017faster} and instance segmentation (e.g., Mask R-CNN~\cite{hemask}) to this canonical task is a promising way to extract object information.
Directly fitting an ellipse on the output mask (i.e., Mask R-CNN in Fig.~\ref{fig:motivation2D}), however, fails to infer the entire object shape.
Intuitively, there are two ways to modify the Mask R-CNN model for predicting ellipses: (1) adding a regression model right after the mask branch; (2) performing regression directly on the features from RoiAlign.
As we show in the ablation study (see Fig.~\ref{fig:ellipseAngleCompare} and Fig.~\ref{fig:expResEllipse}), these two variants both lose the ellipse orientation information, while our Ellipse R-CNN by injecting the learned whole object information (see Fig.~\ref{fig:motivation2D}), achieves the best performance, especially in occluded and clustered scenarios.
Moreover, the detected ellipses can be further exploited for 3D localization and size estimation of such ellipsoid objects (see Fig.~\ref{fig:motivation3D} for cup localization), while the bounding boxes are not reliable in terms of 2D object representation due to insufficient geometric constraints from multiple views~\cite{nicholson2019quadricslam} (see Fig.~\ref{fig:motivation3D} for fruit localization).

Our goal is to accurately detect and represent the elliptical objects in a 2D image as the input for 3D localization, and successfully infer the whole information of each object in an occluded and clustered scenario.
Our approach is motivated by two key ideas.
Common elliptical objects, such as apples, oranges, and peaches, or even cups, can be modeled as ellipsoids in 3D, and their back projections on the 2D image should be ellipses.
The detection mechanism, just as that in the human brain, should be able to retrieve the whole elliptical objects by focusing on their partially visible boundary information so as to handle different occluded patterns effectively.

Our main contributions are twofold:
\begin{itemize}
	\item We propose a robust and compact ellipse regression model that detects each individual elliptical object and parameterizes it as an ellipse.
	The proposed regression method is general and flexible enough to be applicable to any state-of-the-art detection model, in our case, a Mask R-CNN detector~\cite{hemask}.
	For better accuracy, the proposed feature regions before ellipse regression are refined by bounding box estimation and feature padding.
	The correction of the regression strategy is validated cross different synthetic datasets containing well-separated and clustered ellipses, respectively.
	We further analyze the improvement of our ellipse regression in an ablation study using the FDDB dataset~\cite{jain2010fddb}.
	\item For better handling occlusion, we integrate the U-Net~\cite{ronneberger2015u} structure into the detection model to generate decoded feature maps that contain retrieved hidden information.
	We further propose to learn various occluded patterns such that the detection confidence score is computed by generalizing the occlusion information between the visible part and the whole estimated ellipse.
	In ablation experiments, we demonstrate that our approach indeed improves the detection performance compared to the Mask R-CNN baseline and its three variants using both synthetic and real datasets of occluded and clustered objects.
	Moreover, in heavily occlusion settings, our approach achieves the best-reported performance on the datasets.
\end{itemize}

To the best of our knowledge, this paper is the first work developing CNN-based model to detect objects as ellipses and to predict ellipse parameters in one shot directly from the image, and it is the first attempt to handle occlusion from the perspective of ellipse representation.

%% file: tipRelatedWork.tex
\section{Related Work} \label{sec:relatedWork}
Since we develop the Mask R-CNN model as our base object detector to predict ellipse parameters in occluded cases, we review recent work on CNN-based object detectors, 3D object localization, and occlusion handling, respectively.

\emph{CNN-Based Object Detectors.} Recent success in the general object detection tasks on Pascal~\cite{everingham2010pascal}, ImageNet~\cite{krizhevsky2012imagenet}, and MS COCO datasets~\cite{lin2014microsoft}, have been achieved by both single-shot~\cite{redmon2016you, liu2016ssd} and R-CNN~\cite{girshick2015fast, ren2017faster, hemask} architectures.
The single-shot methods formulate object detection as a single-stage regression problem to predict objects extremely fast.
The R-CNN approaches by integrating region proposal and classification, have greatly improved the accuracy, and are currently one of the best performing detection paradigms.

Fruit detection, as a challenging example of elliptical object detection, recently has attracted intensive interests in machine vision and agricultural robotics~\cite{wang2013automated, hung2015feature, roy2016vision}.
Although recent works~\cite{bargoti2017deep, hani2018apple} by tunning head layers of R-CNN models, have presented a good performance on well-separated fruits, the detection accuracy drops significantly as fruits cluster and occlude each other.
The authors in~\cite{hani2018comparative} further report a comparative study of CNN-based models on various apple datasets.
In modern orchards, fruit occlusions happen frequently due to the environmental complexity~\cite{dong2017linear}, especially when they are occluded by neighboring leaves and branches.
It is even more challenging to estimate the 3D size and the 6D pose (position and orientation) of each individual fruit in such occluded and clustered cases.

\emph{Object Localization from 2D Detection.} Recent research has been developed on the size and pose estimation from object detectors by modeling objects as quadrics in 3D~\cite{rubino20183d, nicholson2019quadricslam}, but the whole object information in heavily occluded cases is hardly retrieved from a bounding box that is derived from a little portion of the visible part (see Fig.~\ref{fig:motivation3D}).
While some efforts have been made for 3D fruit localization~\cite{das2015devices, roy2016surveying, roy2018registering, dong2018semantic} by utilizing standard mapping techniques~\cite{wu2013towards, mur2017orb}, none of them could perform object size estimation because of low-resolution 3D reconstructions.
Moreover, all these works represent visible object parts using bounding boxes, which are not appropriate for further estimating the object size and pose due to the spatial ambiguities in rectangle constraints of the ellipsoid.
Specifically, fitting an inscribed ellipse for each bounding box~\cite{rubino20183d} is not reasonable, since there could be infinite solutions for ellipse parameters that all satisfy the bounding box constraints~\cite{nicholson2019quadricslam} even from multiple views.
We thus propose ellipse representation by developing the Mask R-CNN architecture as the baseline, and compare the detection accuracy in terms of masks generated from ellipse parameters.


\emph{Occlusion Handling.} One of the most common applications that apply occlusion handling strategy is pedestrian detection.
The part-based methods~\cite{mathias2013handling} propose to learn a set of specific manually designed occlusion patterns, in which either hand-crafted features~\cite{enzweiler2010multi} or pre-defined semantic parts~\cite{tian2015deep} are employed.
The drawback is that each occlusion pattern detector is learned separately, and it makes the whole procedure complex and hard to train.
In contrast, our approach does not require pre-defining occlusion patterns.
Moreover, pedestrians are shown vertically in the image and represented by bounding boxes, whereas the orientation of elliptical objects when described as ellipses cannot be predetermined and thus increases the complexity of learning patterns.
We incorporate estimated ellipse parameters to learn a continuous vector that serves as a reference to generalize different learned patterns.
The recent work~\cite{sundermeyer2018implicit} learns implicit information to infer the 6D pose of the object by using an autoencoder structure.
In contrast, we integrate the U-Net structure to retrieve occluded information in decoded feature maps.


%% file: tipOverviewBbox.tex
\section{Overview of Bounding-Box Regression} \label{sec:bboxOverview}
\begin{figure}[t]
	\centering
	\includegraphics[width=0.9\columnwidth]{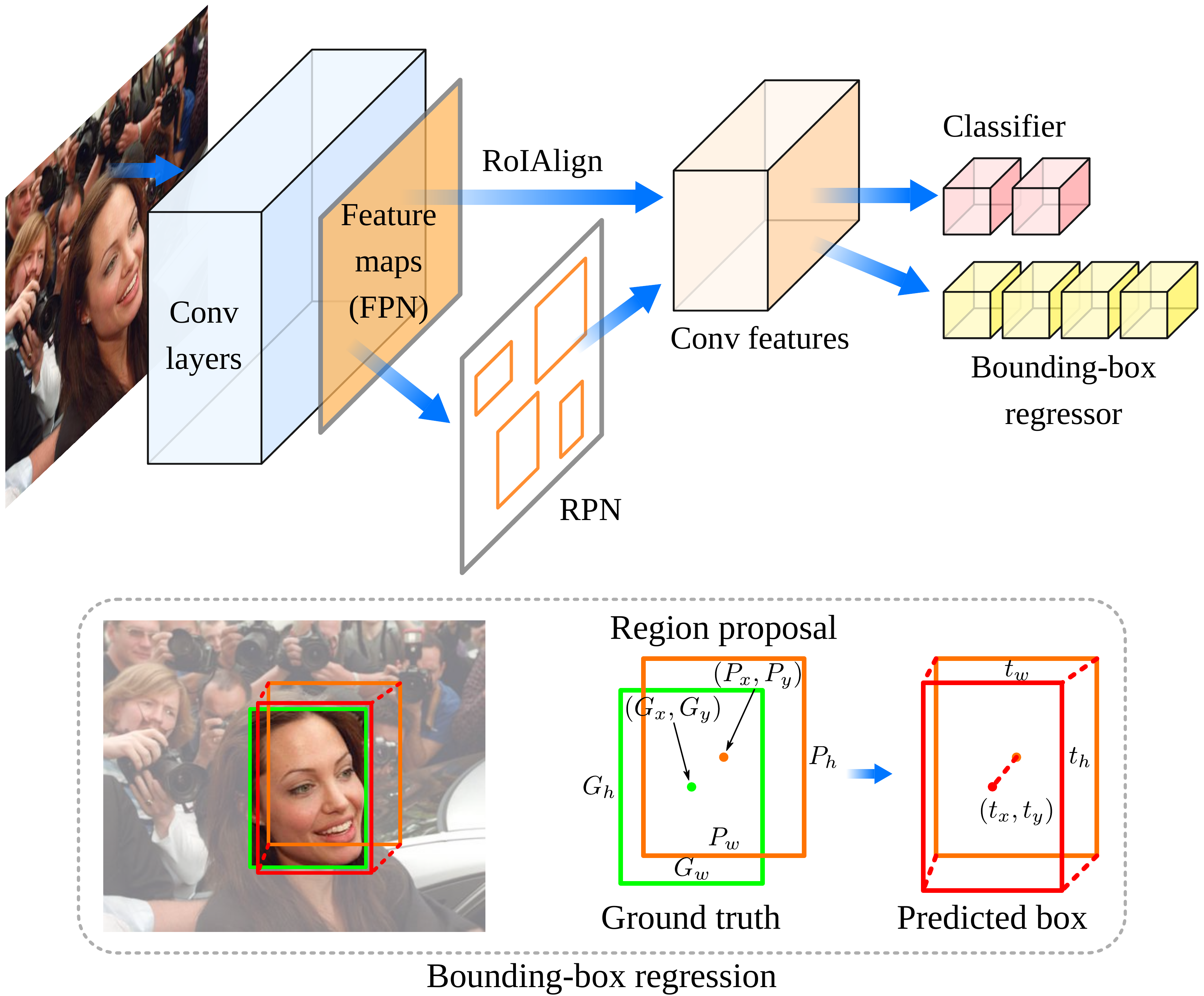}
	\caption{Overview of bounding-box regression in the R-CNN methods (e.g., Mask R-CNN~\cite{hemask}).}
	\label{fig:bboxRegression}
\end{figure}

Most single-stage and R-CNN based networks formulate the object detection as a regression problem, which outputs a class-specific bounding box for each prediction~\cite{redmon2016you, liu2016ssd, redmon2017yolo9000, girshick2015fast, ren2017faster, hemask}.
In our case (see Fig.~\ref{fig:bboxRegression}), the input to the R-CNN models is a set of $N$ training pairs $\{({}^iP, {}^iG)\}_{i = 1, ..., N}$, where ${}^iP = ({}^iP_x, {}^iP_y, {}^iP_w, {}^iP_h)$ denotes the pixel coordinates of the center of a region proposal ${}^iP$ together with its weight and height in pixels.
The ground-truth (GT) bounding box ${}^iG$ corresponding to ${}^iP$ is defined in the same way as ${}^iG = ({}^iG_x, {}^iG_y, {}^iG_w, {}^iG_h)$.
For example, in Mask R-CNN, proposals are generated by a region proposal network (RPN)~\cite{ren2017faster} after the input image going through the base net (i.e., ResNet-101~\cite{he2016deep}).
The feature maps for each proposal are cropped from the top convolutional feature maps generated by feature pyramid network (FPN)~\cite{lin2017feature} according to different proposal scales.
The following RoiAlign layer~\cite{hemask} reshapes the cropped features to produce feature maps of the same size per proposal for classification and bounding-box regression.

The regression goal is to learn a transformation that maps a proposed region ${}^iP$ to a ground-truth box ${}^iG$.
Instead of directly predicting the absolute coordinates and sizes of a bounding box, the model learns to estimate the relative offset parameters $t_x$, $t_y$, $t_w$ and $t_h$ to describe how different the proposal $P$ is compared to the ground-truth $G$ (we drop the superscript $i$ for simplicity):
\begin{equation} \label{eq:bboxReg}
\begin{aligned}
& t_x = (P^{\prime}_x - P_x) / P_w , \quad t_y = (P^{\prime}_y - P_y) / P_h , \\
& t_w = \log (P^{\prime}_w / P_w) , \quad t_h = \log (P^{\prime}_h / P_h) , \\
& t^{\ast}_x = (G_x - P_x) / P_w , \quad t^{\ast}_y = (G_y - P_y) / P_h , \\
& t^{\ast}_w = \log (G_w / P_w) , \quad t^{\ast}_h = \log (G_h / P_h) ,
\end{aligned}
\end{equation}
where $t^{\ast}$ is the regression target, and $P^{\prime}$ denotes the predicted box that is recovered from $t$.
For the object likelihood (of $C$ classes), the model considers the background (i.e., absence of objects) as another class, and predicts the confidence scores of $C+1$ classes.
Specifically, for fruit or face detection, there exists only one class of interest ($C=1$) such that only two values are necessary for the output per proposal with the class determined by the highest one.

There are two key benefits of predicting relative offset parameters in Eq.~\eqref{eq:bboxReg} for accurate bounding-box regression:
\begin{itemize}
	\item All four parameters of each bounding box are normalized such that the objects even with hugely different sizes contribute equally to the total regression loss, which also means that the loss is unaffected by the image size.
	\item The normalization guarantees that all predicted values are close to zero (with small magnitudes) when the proposed region is near the ground-truth box, which stabilizes the training procedure without outputting unbounded values.
\end{itemize}

%% file: tipEllipseRegression.tex
\section{Proposed Ellipse Regressor} \label{sec:ellipseRegression}
Our key idea for ellipse regression is to infer relative offset parameters directly from visible parts so as to maintain the two key benefits described in Sec.~\ref{sec:bboxOverview}.
By further leaning occluded patterns, the confidence score of visible parts of an occluded object is leveraged from incorporated information of its estimated ellipse.
Since Mask R-CNN~\cite{hemask} obtains the state-of-the-art results in general object detection and instance segmentation, we exploit its base model as our front-end network (see Fig.~\ref{fig:motivation2D}).

\begin{figure}[t]
	\centering
	\includegraphics[width=0.99\columnwidth]{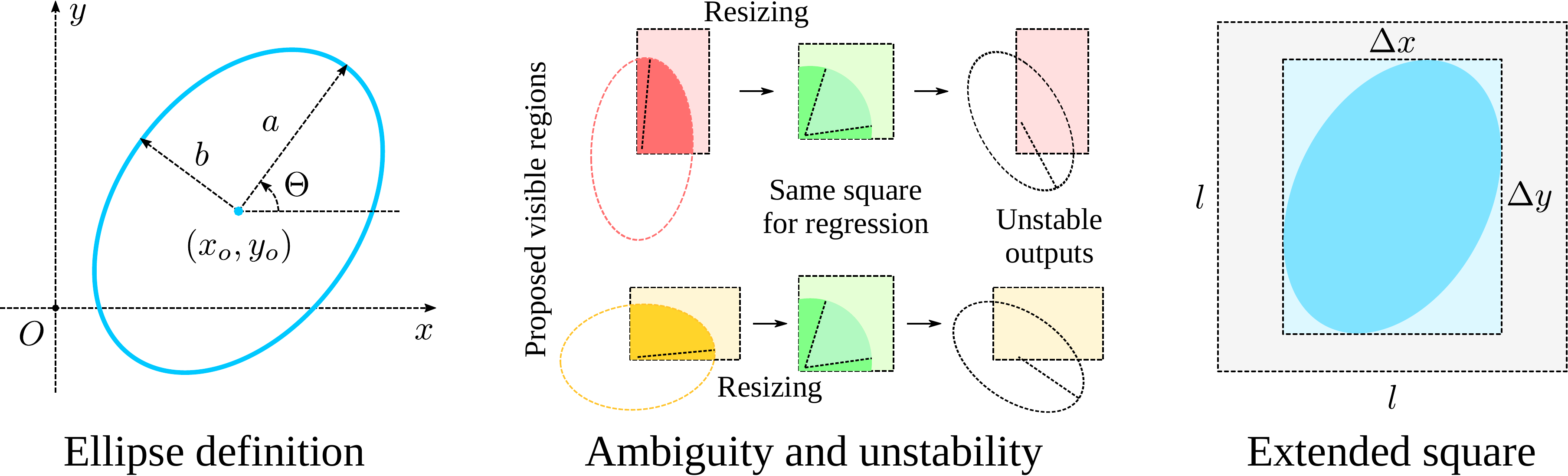}
	\caption{Left: Definition of parameters for a general ellipse. Right: Our extended feature square of an ellipse region. Middle: Directly resizing proposed ellipse regions (red and yellow) as squares (green) has two drawbacks: (1) proposed visible regions (from RPN~\cite{ren2017faster}) of different ellipses may end up with the same resized feature region, which brings orientation ambiguity; (2) original ellipses are distorted and resized close to circular shapes, where the information of orientation and two axes is not salient for regression (error-prone). If the resizing direction (or affine transformation) is tracked, the orientation in (1) can be uniquely determined by the ellipse implicit equation~\cite{young2010precalculus} that is transformed from the resized square. However, the equation coefficients are affected by the predicted axes and orientation in (2) with high uncertainties, which makes the final orientation output unstable. We quantitatively show the unstability in Fig.~\ref{fig:ellipseAngleCompare}.}
	\label{fig:ellipseDefinition}
\end{figure}

\subsection{Formulation of Ellipse Feature Regions} \label{subsec:featureExtension}
In geometry, a general ellipse oriented arbitrarily (see Fig.~\ref{fig:ellipseDefinition}) can be defined by its five parameters: center coordinates $(x_o, y_o)$, semi-major and semi-minor axes $a$, $b$ ($a \ge b$), and rotation angle $\Theta$ (from the positive horizontal axis to the major axis of the ellipse).
The canonical form of the general ellipse~\cite{larson2014precalculus} is obtained as follows
\begin{equation} \label{eq:ellipseForm}
\begin{gathered}
\dfrac{(x^{\prime}\cos \Theta + y^{\prime}\sin \Theta)^2}{a^2} + \dfrac{(-x^{\prime}\sin \Theta + y^{\prime}\cos \Theta)^2}{b^2} = 1 , \\
x^{\prime} = x - x_o , \quad y^{\prime} = y - y_o , 
\end{gathered}
\end{equation}
where the ellipse orientation is $\Theta \in (-\pi/2, \pi/2]$.
We aim to train a regressor for predicting all five ellipse parameters, given a set of $N$ training pairs $\{({}^iP, {}^iE)\}_{i = 1, ..., N}$ as the input to the ellipse regressor, where ${}^iE$ is the ground-truth ellipse characterized by $({}^iE_x, {}^iE_y, {}^iE_a, {}^iE_b, {}^iE_{\Theta})$ and ${}^iP$ is denoted in the same way as in Sec.~\ref{sec:bboxOverview}.
This can be thought of as ellipse regression from a proposed feature region to a nearby ground-truth ellipse.

\begin{figure}[t]
	\centering
	\includegraphics[width=0.99\columnwidth]{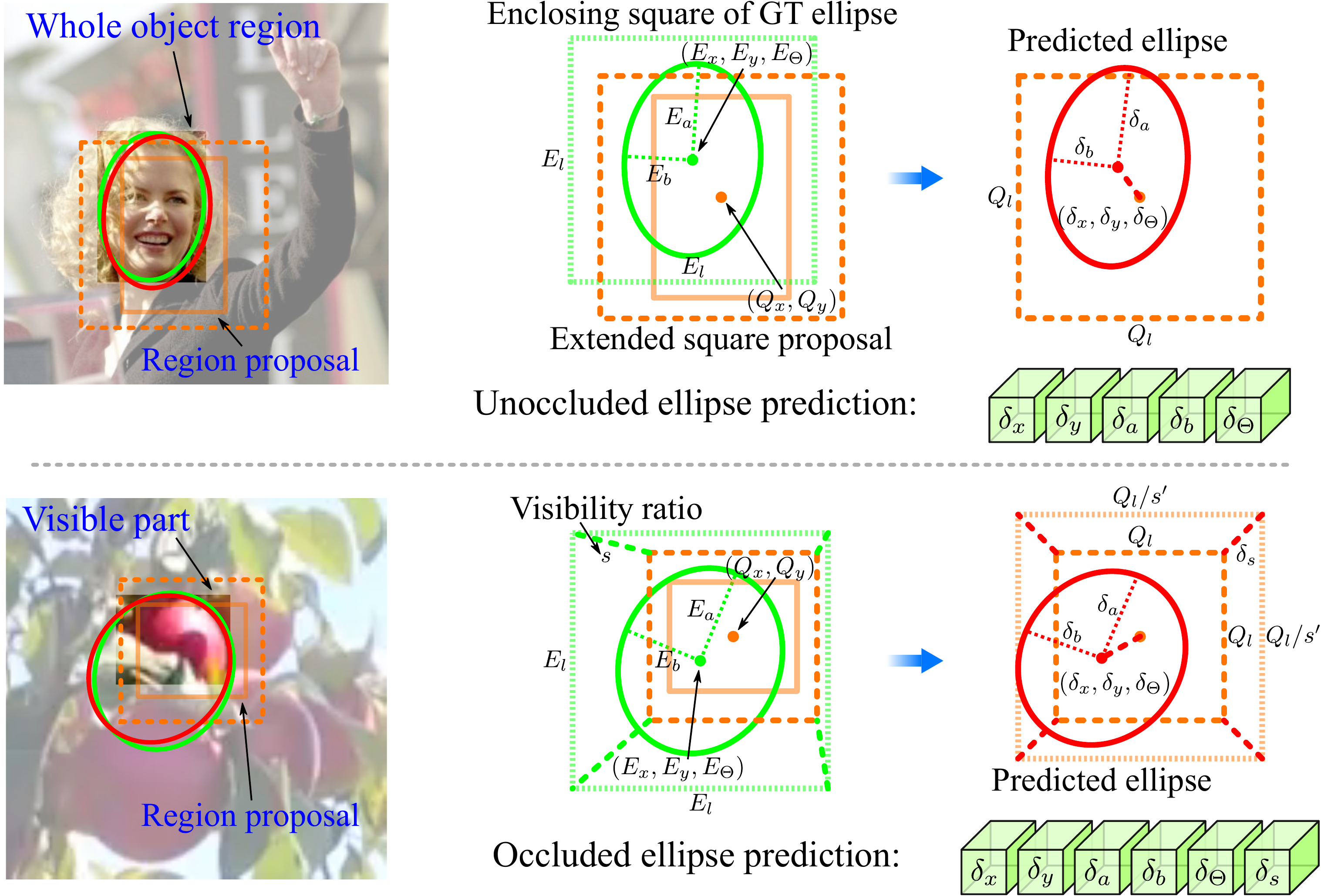}
	\caption{Overview of proposed ellipse regressor developed from unoccluded case (upper) to occluded case (lower). The regressor handles both cases by predicting one more offset $\delta_s$ that estimates the visibility ratio $s$ between the extended square $Q$ of the proposed visible part and the enclosing square of the ground-truth (GT) ellipse $E$.}
	\label{fig:ellipseRegression}
\end{figure}

However, the strategy of bounding-box regression cannot be directly applied to ellipse regression.
The major challenge comes from how to accurately keep the ellipse orientation information in each feature region $^iP$ before the regression stage.
For example, the RoiAlign layer~\cite{hemask} in the state-of-the-art methods resizes the rectangular proposed regions of various shapes as squares of a fixed size, but this distorts the features maps and makes the prediction of the original ellipse orientation information unstable (see Fig.~\ref{fig:ellipseDefinition} for more details).

We therefore propose, before the resizing operation, to extend each rectangular feature proposal ${}^iP$ as a squared region ${}^iQ$, whose length ${}^iQ_l$ only depends on the axes sizes of the ellipse to be predicted.
The length ${}^iQ_l$ of the extended square region ${}^iQ$ is derived as follows.
In Eq.~\eqref{eq:ellipseForm}, we take the derivative of $y$ with respect to $x$:
\begin{equation} \label{eq:ellipseDerive}
\scalebox{0.95}{$
	\dfrac{\partial y}{\partial x} = - \dfrac{a^2x^{\prime}\sin^2 \Theta - y^{\prime}(a^2 - b^2)\sin\Theta\cos \Theta + b^2x^{\prime}\cos^2 \Theta}{a^2x^{\prime}\cos^2 \Theta - x^{\prime}(a^2 - b^2)\sin\Theta\cos \Theta + b^2y^{\prime}\sin^2 \Theta}
	$} . \\
\end{equation}
To determine the axis-aligned bounding box for the ellipse,
we equate the numerator and denominator of Eq.~\eqref{eq:ellipseDerive} to zero separately, since zero numerator and denominator correspond to horizontal and vertical tangents of the ellipse, respectively.
The bounding-box length along each axis is solved as:
\begin{equation} \label{eq:ellipseLimits}
\begin{cases}
\Delta x = 2 \sqrt{a^2\cos^2\Theta + b^2\sin^2\Theta} \\
\Delta y = 2 \sqrt{a^2\sin^2\Theta + b^2\cos^2\Theta}
\end{cases} .
\end{equation}
We further create an extended square enclosing the ellipse bounding box, whose diagonal length is defined as the square length $l = \sqrt{{\Delta x}^2 + {\Delta y}^2} = 2\sqrt{a^2 + b^2}$.

\begin{figure*}[!t]
	\centering
	\includegraphics[width=0.99\textwidth]{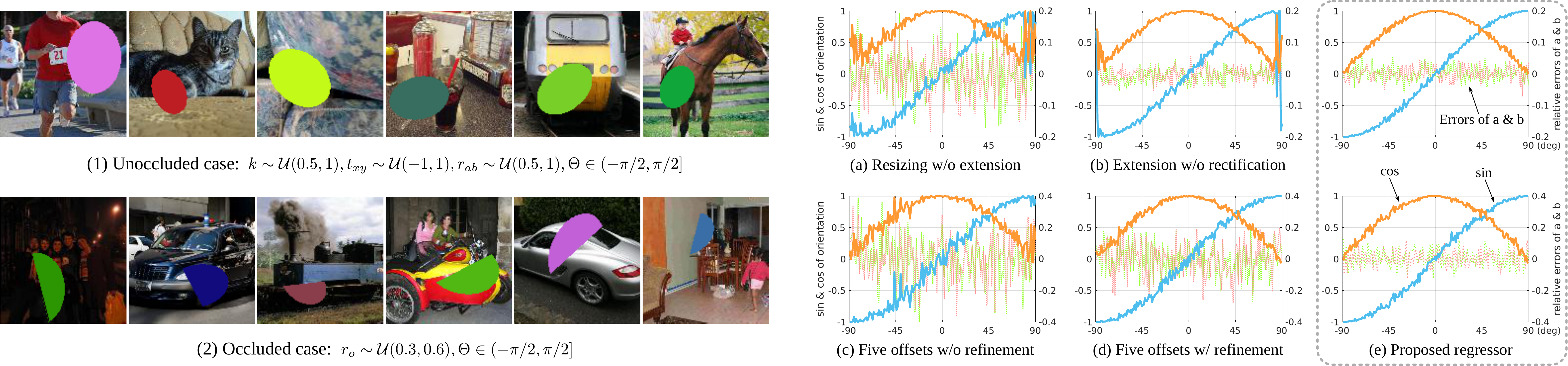}
	\caption{Comparison between different ellipse regression strategies training and testing on unoccluded (1) and occluded (2) ellipses, respectively. Left: Synthetic images (with the background filled by the Pascal dataset~\cite{everingham2010pascal} to add interference) where ellipses are randomly generated from the distribution~\cite{mattheydsprites} parameterized by size scale $k$, center translation $t_{xy}$, axes ratio $r_{ab}$, and visibility ratio $r_o$. Right: Sine (blue) and cosine (orange) values of predicted ellipse angle $\Theta$ and relative errors of two estimated axes $a$ (green) and $b$ (pink). Our proposed regressor (e) accurately predicts ellipse parameters and handles both unoccluded and occluded cases, outperforming the other four variants: (a) directly resizing rectangular feature regions as squares, (b) extending feature regions in Sec.~\ref{subsec:featureExtension} without angle rectification in Eq.~\eqref{eq:ellipseRegLoss}, (c) predicting five offsets in Eq.~\eqref{eq:ellipseReg} without feature region refinement in Sec.~\ref{subsec:featureRefinement}, and (d) predicting five offsets from the refined feature region.}
	\label{fig:ellipseAngleCompare}
\end{figure*}

Thus, given a proposal ${}^iP$ closely bounding the ellipse ${}^iE$, we extend it as the square ${}^iQ$ sharing the same center $({}^iP_x, {}^iP_y)$ with its length as ${}^iQ_l = \sqrt{{{}^iP_w}^2 + {{}^iP_h}^2}$.
Besides no distortion of ellipse orientation, another advantage of the extended feature region ${}^iQ$ is that its size ${}^iQ_l$ is still proportional to the ellipse size ($a$ and $b$) but independent on the ellipse angle $\Theta$.
\textit{It implies that even ellipses of the same size ($a$ and $b$) but with different orientation angles will contribute equally to the regression loss (otherwise the different sizes of their axis-aligned bounding boxes will weight inconsistently in the loss and make the regression model sensitive to the ellipse angle).}
Our extending strategy also addresses the issue in direct resizing methods (see Fig.~\ref{fig:ellipseDefinition}): since the prediction of $\Theta$ is coupled with the ellipse region shape (see $\Delta x$ and $\Delta y$ in Eq.~\eqref{eq:ellipseLimits}), the resizing step complicates the orientation learning process.
The extended feature region ${}^iQ$ thus serves as a stable reference to accurately predict ellipse offset parameters.

\subsection{Ellipse Offsets Prediction}
Given a squared feature region ${}^iQ = ({}^iQ_x, {}^iQ_y, {}^iQ_l)$ extended from a proposal $P^i$, our goal is to learn regressing features within ${}^iQ$ to a set of relative offset parameters between ${}^iQ$ and a ground-truth ellipse ${}^iE$.
We start from predicting elliptical objects without occlusion, and propose stable offset parameters to handle occluded cases.

\subsubsection{Unoccluded Ellipse Prediction}
For well-separated objects, we parameterize the regression in terms of five outputs $\delta_x$, $\delta_y$, $\delta_a$, $\delta_b$ and $\delta_{\Theta}$ (superscript $i$ is dropped for simplicity):
\begin{equation} \label{eq:ellipseReg}
\begin{aligned}
& \delta_x = (E^{\prime}_x - Q_x) / Q_l , \; \delta_y = (E^{\prime}_y - Q_y) / Q_l , \; \delta_{\Theta} = E^{\prime}_{\Theta} / \pi , \\
& \delta_a = \log (2E^{\prime}_a / Q_l) , \; \delta_b = \log (2E^{\prime}_b / Q_l) , \\
& \delta^{\ast}_x = (E_x - Q_x) / Q_l , \; \delta^{\ast}_y = (E_y - Q_y) / Q_l , \; \delta^{\ast}_{\Theta} = E_{\Theta} / \pi , \\
& \delta^{\ast}_a = \log (2E_a / Q_l) , \; \delta^{\ast}_b = \log (2E_b / Q_l) ,
\end{aligned}
\end{equation}
where the range of the GT ellipse angle is $E_{\Theta} \in (-\pi/2, \pi/2]$, $\delta^{\ast}$ is the ellipse regression target, and $E^{\prime}$ is the predicted ellipse calculated from $\delta$.
$(\delta_x, \delta_y)$ specifies the scale-invariant translation of the center of $Q$ to $E^{\prime}$, while $\delta_a$and $\delta_b$ specify the log-space translations of the size of $Q$ to semi-major and semi-minor axes of $E^{\prime}$, respectively.
$\delta_{\Theta}$ is the prediction of the normalized orientation of $E^{\prime}$.
In such an unoccluded case ($Q_l \gg 0$), the predicted offset values $\delta$ are all bounded when the proposed region $P$ ($\rightarrow Q$) is located close to the ground-truth ellipse $E$ (see Fig.~\ref{fig:ellipseRegression}).

\subsubsection{Occluded Ellipse Prediction}
For occluded object detection, training RPN~\cite{ren2017faster} to propose regions of visible parts (instead of whole object regions) highly reduces false positives as shown in Sec.~\ref{sec:experiments}.
We infer the whole elliptical object from its visible part through ellipse regression.

However, as the visible region goes small ($Q_l \rightarrow 0$) and locates around the object boundary ($|E_x - Q_x| \gg 0$, and $|E_y - Q_y| \gg 0$), the target values to learn ($\delta^{\ast}_x$, $\delta^{\ast}_y$, $\delta^{\ast}_a$, and $\delta^{\ast}_b$) in Eq.~\eqref{eq:ellipseReg} have unboundedly large magnitudes, which unstabilizes the training process.
We thus propose to predict one more offset parameter $\delta_s$ for the scale $s$ (see Fig.~\ref{fig:ellipseRegression}):
\begin{equation} \label{eq:ellipseRegScale}
\begin{aligned}
& \delta_x = s^{\prime}(E^{\prime}_x - Q_x) / Q_l , \quad \delta_y = s^{\prime}(E^{\prime}_y - Q_y) / Q_l , \\
& \delta_a = \log (2s^{\prime}E^{\prime}_a / Q_l) , \quad \delta_b = \log (2s^{\prime}E^{\prime}_b / Q_l) , \\
& \delta_s = \log \left((s^{\prime} + 1) / 2\right) , \quad \delta_{\Theta} = E^{\prime}_{\Theta} / \pi , \\
& \delta^{\ast}_x = s(E_x - Q_x) / Q_l , \quad \delta^{\ast}_y = s(E_y - Q_y) / Q_l , \\
& \delta^{\ast}_a = \log (2sE_a / Q_l) , \quad \delta^{\ast}_b = \log (2sE_b / Q_l) , \\
& \delta^{\ast}_s = \log \left((s + 1) / 2\right) , \quad \delta^{\ast}_{\Theta} = E_{\Theta} / \pi ,
\end{aligned}
\end{equation}
where $s = Q_l / E_l$, $s \in (0, 1]$ characterizing the visibility ratio calculated between the size $Q_l$ of the extended square $Q$ (from the visible part) and the length $E_l = 2\sqrt{{E_a}^2 + {E_b}^2}$ of the square enclosing the ellipse $E$ (i.e., the whole object region).
By predicting $s^{\prime}$, we transfer the offset reference from the visible part $Q_l$ to the whole object region $Q_l / s^{\prime}$, which guarantees that all predicted values $\delta$ (with the target $\delta^{\ast}$) are bounded even in heavily occluded cases (when the proposed region is near the small visible part).
Specifically, as $s = 1$ and $\delta^{\ast}_s = 0$, Eq.~\eqref{eq:ellipseRegScale} and Eq.~\eqref{eq:ellipseReg} are equivalent, which means that Eq.~\eqref{eq:ellipseRegScale} is a generalized formulation of ellipse offsets prediction that can handle both unoccluded and occluded cases.

After learning such offset parameters $\delta$, we can transform an input extended region $Q$ into a predicted ellipse $E^{\prime}$ by applying the transformation:
\begin{equation} \label{eq:ellipseInference}
\begin{aligned}
& E^{\prime}_x = \tfrac{Q_l}{s^{\prime}} \delta_x + Q_x , \; E^{\prime}_y = \tfrac{Q_l}{s^{\prime}} \delta_y + Q_y , \; s^{\prime} = 2\exp (\delta_s) - 1 , \\
& E^{\prime}_a = \tfrac{Q_l}{2s^{\prime}} \exp (\delta_a) , \; E^{\prime}_b = \tfrac{Q_l}{2s^{\prime}} \exp (\delta_b) , \; \theta^{\prime} = \pi \delta_{\Theta} , \\
& E^{\prime}_{\Theta} =
\begin{cases}
{\rm atan2}\left(\sin\theta^{\prime}, \cos\theta^{\prime}\right)  &\text{if } \cos\theta^{\prime} \ge 0 \\
{\rm atan2}\left(-\sin\theta^{\prime}, -\cos\theta^{\prime}\right)  &\text{if } \cos\theta^{\prime} < 0
\end{cases} ,
\end{aligned}
\end{equation}
where $E^{\prime}_{\Theta}$ is rectified from $\theta^{\prime}$ such that $E^{\prime}_{\Theta} \in (-\pi/2, \pi/2]$.

\subsubsection{Ellipse Regression Loss}
For a proposed region ${}^iP$, we define the regression loss as:
\begin{equation} \label{eq:ellipseRegLossTotal}
{\mathcal{L}_{\text{e}}}^{(i)} =
\begin{cases}
{}^i\mathbb{I}^{\ast} ~ \mathcal{R}\left({}^i\delta_{\star} - {}^i\delta^{\ast}_{\star}\right) & \text{if } \star \in \{x, y, a, b, s\} \\
{}^i\mathbb{I}^{\ast} ~ \mathcal{R}\left(\tfrac{1}{\pi} \rho \left({}^i\delta_{\star}, {}^i\delta^{\ast}_{\star}\right)\right) & \text{if } \star = \Theta
\end{cases} ,
\end{equation}
where ${}^i\mathbb{I}^{\ast} = 1$ indicates that ${}^iP$ is positive (if the intersection-over-union (IoU) overlap with its ground-truth box ${}^iG$ is higher than a ratio~\cite{ren2017faster}), while ${}^i\mathbb{I}^{\ast} = 0$ if ${}^iP$ is non-positive.
$\mathcal{R}$ is the robust loss function (smooth $\text{L}_1$) defined in~\cite{girshick2015fast}, and $\rho$ is the transformation function defined as:
\begin{equation} \label{eq:ellipseRegLoss}
\begin{aligned}
\rho ({}^i\delta_{\Theta}, {}^i\delta^{\ast}_{\Theta}) &=
\begin{cases}
{\rm atan2}\left(\sin{}^i\varphi, \cos{}^i\varphi\right) & \text{if } \cos{}^i\varphi \ge 0 \\
{\rm atan2}\left(-\sin{}^i\varphi, -\cos{}^i\varphi\right) & \text{if } \cos{}^i\varphi < 0
\end{cases} , \\
{}^i\varphi &= \pi ({}^i\delta_{\Theta} - {}^i\delta^{\ast}_{\Theta}) ,
\end{aligned}
\end{equation}
which rectifies the ellipse orientation loss ${}^i\varphi$ of ${}^i\delta_{\Theta}$ compared to ${}^i\delta^{\ast}_{\Theta}$ around critical angles (for example, the angle difference between $\pi/2$ and $-\pi/2$ should be zero rather than $\pi$).

\subsection{Feature Region Refinement} \label{subsec:featureRefinement}

\begin{figure}[t]
	\centering
	\includegraphics[width=0.95\columnwidth]{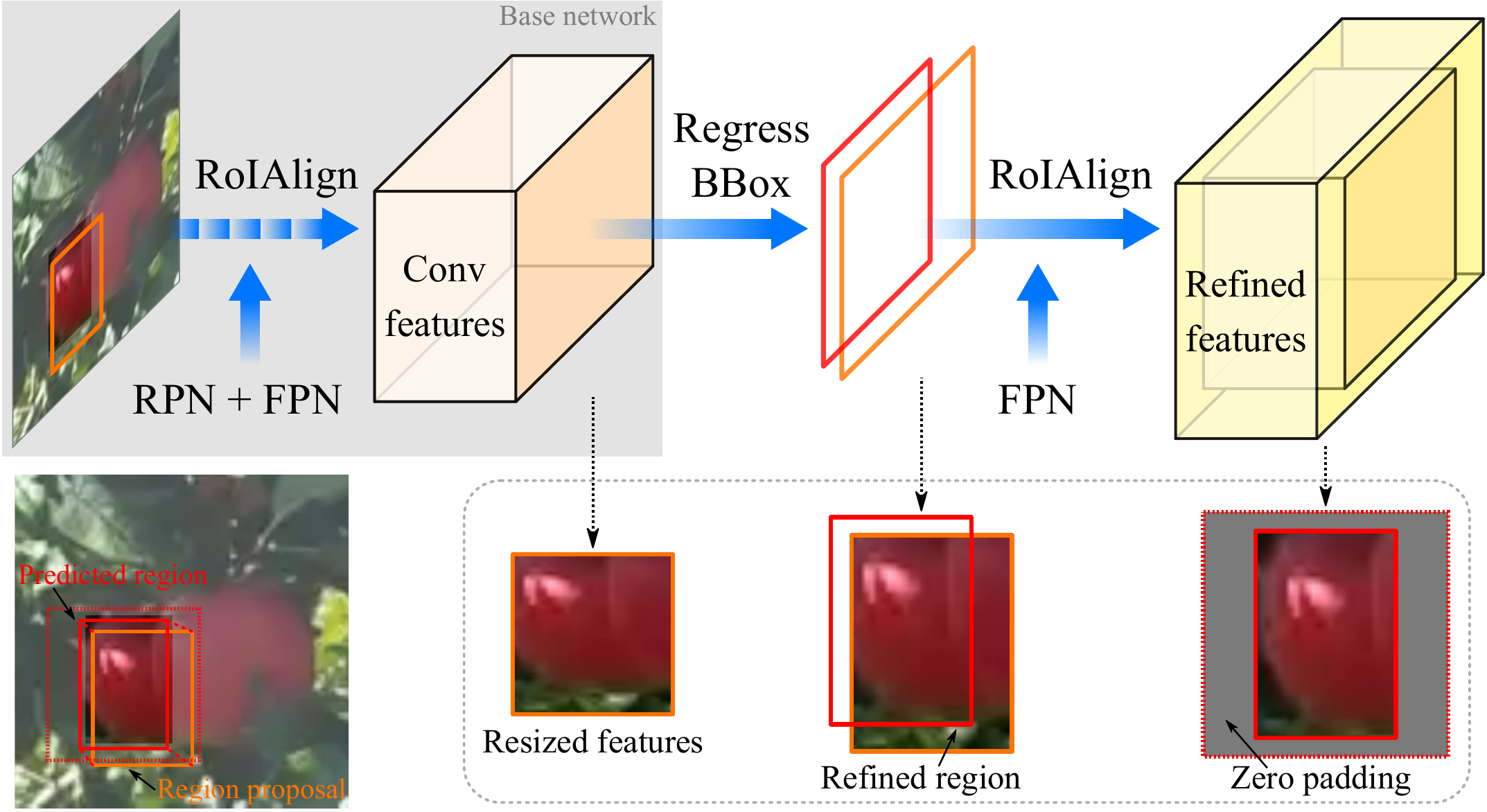}
	\caption{An example of how the feature region refinement is performed based on visible region prediction and zero padding. RPN + FPN: the region proposal network (RPN) and the feature pyramid network (FPN) from the base network of Mask R-CNN (see Fig.~\ref{fig:motivation2D}).}
	\label{fig:featureRefinement}
\end{figure}

Traditional R-CNN based methods generate regression and classification outputs directly from proposed feature regions.
However, relying on only roughly proposed feature maps from RPN maybe risky and error-prone especially to predict ellipse orientation in heavily occluded cases (see Fig.~\ref{fig:ellipseAngleCompare}).
Specifically, there exists a mismatch between a predicted visible region and its feature representation (see Fig.~\ref{fig:featureRefinement}).
Thus, our idea is to perform ellipse regression and classification based on the refined feature region output by a bounding-box regressor.
This strategy alleviates the issue by allowing the model to exploit the features of the exact predicted visible region, which makes the inference output more reliable.

Based on the extended predicted region, the RoiAlign layer re-extracts a small feature map (e.g., 14$\times$14), and accurately aligns the extracted features with the input from FPN.
Features in the extended square but located outside the predicted visible region have a negative effect on predicting accurate ellipse parameters (see large numeric errors in Fig.~\ref{fig:ellipseAngleCompare} (c)).
To reduce the interference of such unrelated features, we perform zero padding on the extended feature area.
Our proposed method is simple: we use floor and ceiling operations to compute the boundaries of the smallest rectangle that encloses the bilinear-interpolated feature map from the predicted region, and pad zeros in the rest area of the extended square.
For example, $\lfloor -\frac{w}{2r} \rfloor$ and $\lceil \frac{w}{2r} \rceil$ are two width limits of the resized rectangle whose center is assumed at $(0, 0)$, where $w$ is the width of the predicted region and $r$ is the resizing factor.
The refined feature region leads to large improvements as we show in Sec~\ref{sec:experiments}.

\subsection{Learning Occlusion Patterns}
\begin{figure}[t]
	\centering
	\includegraphics[width=0.99\columnwidth]{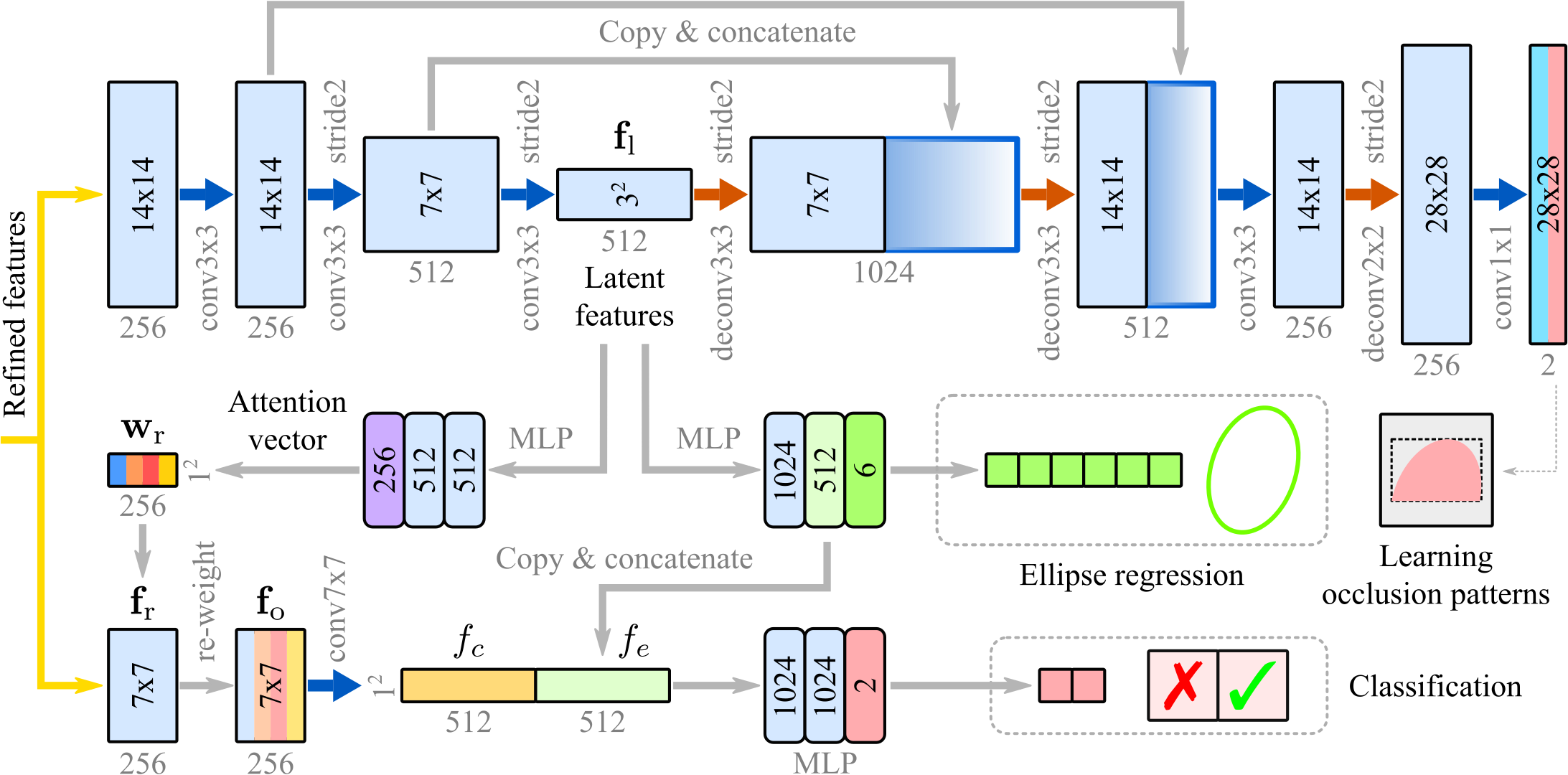}
	\caption{Head architecture of learning occlusion patterns for ellipse regression and classification. The refined features of two different sizes both serve as the input. Numbers in MLP blocks stand for layers sizes. Batch normalization~\cite{ioffe2015batch} is applied for all layers with ReLU~\cite{nair2010rectified} except for the last output layers.}
	\label{fig:learningOcclusion}
\end{figure}

Diverse appearances of occluded objects lead to a large variety of occlusion patterns (see Fig.~\ref{fig:motivation2D}).
Traditional networks are likely to assign a low confidence to an occluded object due to its hidden parts.
Our key idea for occlusion handling is to employ channel-wise attention in refined features by learning different occlusion patterns in one coherent model.
Our model can leverage the prediction confidence of the visible part of an elliptical object based on the inference of the whole ellipse from the occlusion (see Fig.~\ref{fig:learningOcclusion}).

\begin{figure*}[!t]
	\centering
	\includegraphics[width=0.99\textwidth]{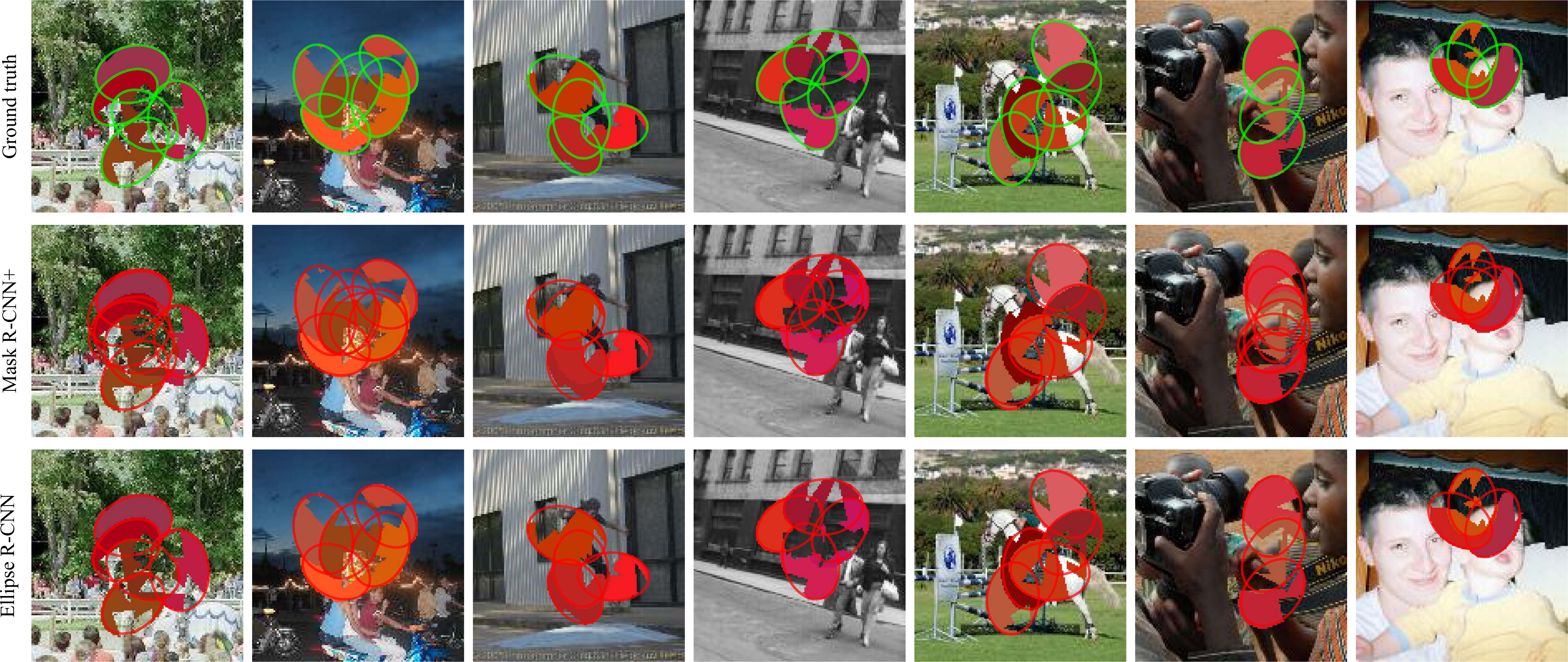}
	\caption{Qualitative results from our Ellipse R-CNN and Mask R-CNN+ on the SOE dataset. The GT ellipse parameters are perfectly generated based on the synthetic ellipse shapes to validate the correctness of the ellipse regression. The Mask R-CNN+ trained on the whole ellipse regions outputs many false positives in heavily occluded cases.}
	\label{fig:expResEllipse}
\end{figure*}

\begin{table*}[thbp!]
	\caption{Overall performance on the SOE dataset. Higher AP is better while lower is better for MR. $\textbf{R}$: ellipse regression with feature refinement. $\textbf{O}$: learning occlusion patterns. $\textbf{A}$: re-weighting refined features by attention. *: $f_e$ is not concatenated with $f_c$. \newline $\text{AP}_{\star}$ and $\text{MR}_{\star}$: the ellipse IoU level starts from 0.75 to 0.95 with an interval 0.05. $\text{AP}_{\star}^{\Theta}$: the angle error decreases from $10^{\circ}$ to $2^{\circ}$ with an interval $2^{\circ}$. $\text{MR}_{\star}^{\Theta}$: the error is from $5^{\circ}$ to $1^{\circ}$ with an interval $1^{\circ}$. The default ellipse IoU for $\text{AP}^{\Theta}$ and $\text{MR}^{\Theta}$ is 0.75.} \label{tab:ellipse}
	\begin{center}
		\begin{tabular}{l|M{.22cm} M{.22cm} M{.22cm}|M{.7cm} M{.7cm} M{.7cm}|M{.7cm} M{.7cm} M{.7cm}|M{.7cm} M{.7cm}|M{.7cm} M{.7cm}}
			\toprule
			Methods & \textbf{R} & \textbf{O} & \textbf{A} & $\text{AP}_{\star}$ & $\text{AP}_{75}$ & $\text{AP}_{85}$ & $\text{MR}_{\star}$ & $\text{MR}_{75}$ & $\text{MR}_{85}$ & $\text{AP}_{\star}^{\Theta}$ & $\text{AP}_{10}^{\Theta}$ & $\text{MR}_{\star}^{\Theta}$ & $\text{MR}_{4}^{\Theta}$ \\
			\bottomrule\toprule
			Mask R-CNN+ & -- & -- & -- & 30.4 & 66.8 & 27.1 & 74.5 & 40.0 & 79.5 & 23.7 & 39.1 & 55.2 & 47.2 \\
			Ellipse R-CNN- &  &  &  & 37.1 & 73.2 & 36.7 & 67.9 & 30.1 & 68.0 & 34.2 & 50.6 & 44.0 & 38.3 \\
			Ellipse R-CNN- & \checkmark &  &  & 44.9 & 83.1 & 48.1 & 62.4 & 23.2 & 69.9 & 43.6 & 65.2 & 35.9 & 27.1 \\
			\midrule
			DeepParts+~\cite{tian2015deep} & \checkmark & \checkmark &  & 47.5 & 85.7 & 53.6 & 58.3 & 17.9 & 54.7 & 47.7 & 69.5 & 31.6 & 23.7 \\
			SENet+~\cite{hu2018squeeze} & \checkmark &  & \checkmark & 48.0 & 86.5 & 53.9 & 57.7 & 16.8 & 55.0 & 48.1 & 71.1 & 30.5 & 22.2 \\
			Ellipse R-CNN* & \checkmark & \checkmark & \checkmark & 49.2 & 88.1 & 56.0 & 55.2 & 15.5 & 53.0 & 49.9 & 73.7 & 27.3 & 20.9 \\
			Ellipse R-CNN & \checkmark & \checkmark & \checkmark & \textbf{51.5} & \textbf{89.9} & \textbf{58.9} & \textbf{53.8} & \textbf{13.3} & \textbf{50.9} & \textbf{52.3} & \textbf{76.8} & \textbf{24.7} & \textbf{18.5} \\
			\bottomrule
		\end{tabular}
	\end{center}
\end{table*}

\subsubsection{Occluded Ellipse Patterns}
Given the refined features of a predicted visible region, we exploit a U-Net~\cite{ronneberger2015u} structure to learn the occluded ellipse shape within the extended square (see Fig.~\ref{fig:learningOcclusion}).
The ground truth of the occluded ellipse shape is generated as follows.
For an occluded object, we identify a bounding box of the visible part with its ellipse parameters. The GT whole ellipse generated is then cropped and resized by a predicted visible region, and put centered in the extended square. The GT visible ellipse is thus obtained without being occluded by other nearby obstacles.
Unlike previous work~\cite{tian2015deep, zhou2017multi}, our method does not relies on any particular discrete set of occlusion patterns or any external classifier for guidance, and thus can be trained in an end-to-end manner.

By learning occluded ellipse patterns, the low-dimensional latent features $\mathbf{f}_\text{l}$ encode both partial visibility and ellipse shape information~\cite{sundermeyer2018implicit}.
Therefore, we perform ellipse regression directly from the latent features.
The ellipse offsets are obtained via a multilayer perceptron (MLP)~\cite{bishop1995neural} (see Fig.~\ref{fig:learningOcclusion}).

\subsubsection{Visible Part Attention}
Many recent works~\cite{bau2017network, gonzalez2018semantic} find that convolution filters of different feature channels respond to their specific high-level concepts, which are associated with different semantic parts.
To leverage the detection confidence in occluded cases, our intuition is to allow the network to decide how much each channel should contribute in the refined features $\mathbf{f}_\text{r}$.
Specifically, the channels representing the visible parts should be weighted more, while the occluded parts be weighted less.
We thus re-weight the refined features as $\mathbf{f}_\text{o}$:
\begin{equation}
\mathbf{f}_\text{o} = \mathbf{w}_\text{r}^{\top} \mathbf{f}_\text{r}, \quad \mathbf{f}_\text{r} = [f_1, f_2, ..., f_H]^{\top},
\end{equation}
where $\mathbf{w}_\text{r}$ is the attention weighting vector regressed from the latent features $\mathbf{f}_\text{l}$ (learned partial visibility) by an MLP, and $H$ is the total number of channels (e.g., 256).
The re-weighted features $\mathbf{f}_\text{o}$ is further regressed as a feature $f_c$ for classification.

Various ellipse orientations may increase the learning complexity of occlusion patterns.
To compensate for the orientation effect, we propose to concatenate the feature $f_c$ with a latent feature $f_\text{e}$ (used for ellipse regression in Fig.~\ref{fig:learningOcclusion}) to incorporate both partial visibility and whole ellipse information.
The concatenated feature $[f_\text{c}, f_\text{e}]$ thus learns various occlusion patterns, and passes through the classification head to output the final confidence scores.

\subsubsection{Training Objective}
The R-CNN based models have two types of losses: RPN loss $\mathcal{L}_\text{RPN}$ and head loss~\cite{ren2017faster} (composed of classification loss $\mathcal{L}_{\text{cls}}$ and regression loss $\mathcal{L}_{\text{reg}}$).
We redefine $\mathcal{L}_{\text{reg}}$ as the sum of the loss of the feature region refinement and the ellipse regression loss $\mathcal{L}_{\text{e}}$.
On top of that, our occlusion handling introduces one additional loss $\mathcal{L}_\text{occl}$ defined as the average binary cross-entropy loss.
The loss function of the whole system can be written as follows:
\begin{equation}
\mathcal{L} = \mathcal{L}_\text{RPN} + \dfrac{1}{N}\sum_{i} \left( {\mathcal{L}_{\text{cls}}}^{(i)} + {^{i}p^{\ast}} \left({\mathcal{L}_{\text{reg}}}^{(i)} + {\mathcal{L}_{\text{occl}}}^{(i)}\right) \right) ,
\end{equation}
where the loss $\mathcal{L}_{\text{cls}}$ is over two classes (object vs. background), and the GT label ${^{i}p^{\ast}}$ is 1 if feature region $i$ (in total $N$ regions) is positive (as an object) otherwise ${^{i}p^{\ast}}$ is 0.

%% file: tipExperiments.tex
\begin{figure*}[!t]
	\centering
	\includegraphics[width=0.99\textwidth]{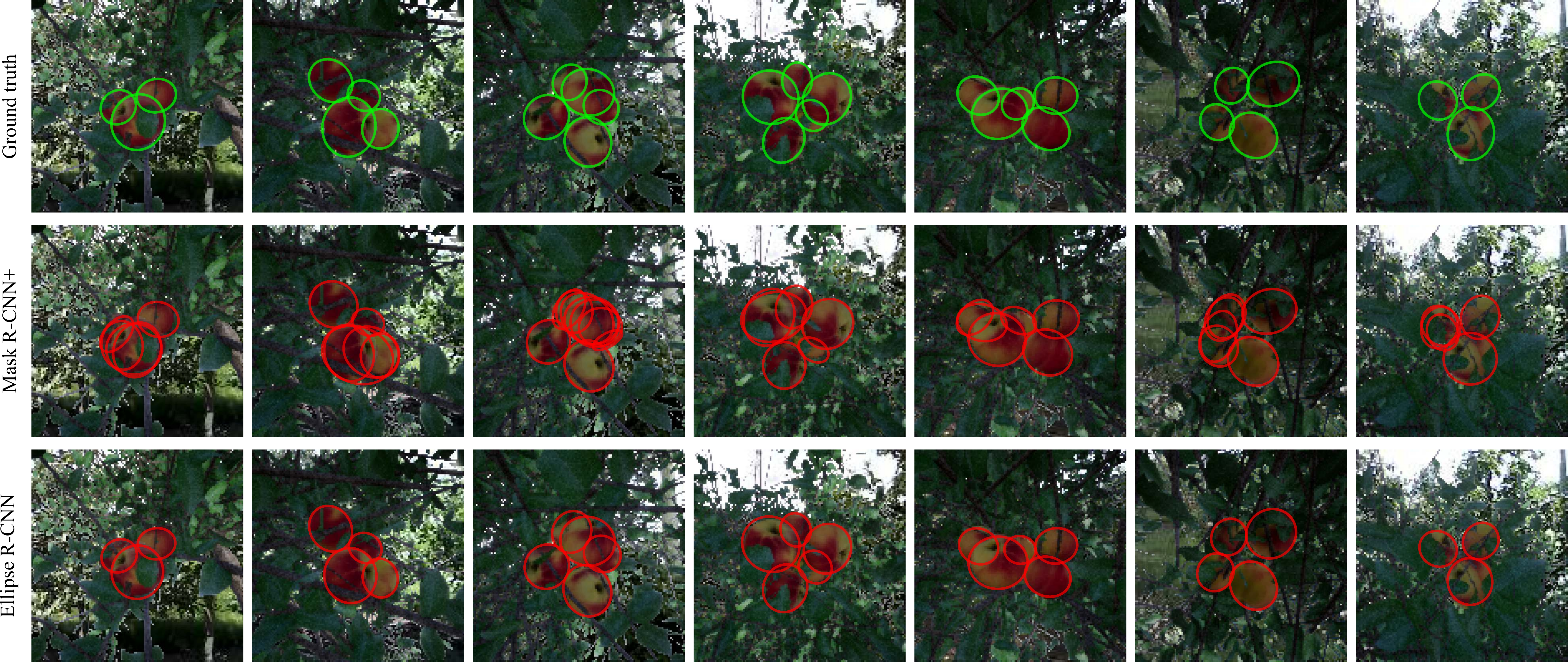}
	\caption{Qualitative results from our Ellipse R-CNN and Mask R-CNN+ on the SOF dataset. The GT ellipse parameters are perfectly generated based on the projective geometry of 3D fruit models. The Mask R-CNN+ trained on the whole object regions outputs many false positives in heavily occluded cases.}
	\label{fig:expResUE}
\end{figure*}

\begin{table*}
	\caption{Overall performance on the SOF dataset. $\text{AP}_{\star}$ and $\text{MR}_{\star}$: the ellipse IoU level starts from 0.7 to 0.9 with an interval 0.05. \newline $\text{AP}_{\star}^{\Theta}$: the angle error decreases from $45^{\circ}$ to $5^{\circ}$ with an interval $10^{\circ}$. The default ellipse IoU for $\text{AP}^{\Theta}$ is 0.7.} \label{tab:ue}
	\begin{center}
		\begin{tabular}{l|M{.22cm} M{.22cm} M{.22cm}|M{.7cm} M{.7cm} M{.7cm}|M{.7cm} M{.7cm} M{.7cm}|M{.7cm} M{.7cm} M{.7cm}}
			\toprule
			Methods & \textbf{R} & \textbf{O} & \textbf{A} & $\text{AP}_{\star}$ & $\text{AP}_{70}$ & $\text{AP}_{80}$ & $\text{MR}_{\star}$ & $\text{MR}_{70}$ & $\text{MR}_{80}$ & $\text{AP}_{\star}^{\Theta}$ & $\text{AP}_{45}^{\Theta}$ & $\text{AP}_{25}^{\Theta}$ \\
			\bottomrule\toprule
			Mask R-CNN+ & -- & -- & -- & 25.7 & 59.7 & 20.3 & 78.3 & 46.6 & 84.1 & 25.6 & 43.0 & 27.6 \\
			Ellipse R-CNN- &  &  &  & 30.3 & 66.2 & 25.1 & 74.3 & 38.4 & 79.5 & 34.9 & 54.9 & 39.7 \\
			Ellipse R-CNN- & \checkmark &  &  & 35.4 & 73.8 & 33.9 & 69.5 & 27.2 & 74.3 & 46.1 & 64.3 & 53.7 \\
			\midrule
			DeepParts+~\cite{tian2015deep} & \checkmark & \checkmark &  & 38.3 & 79.6 & 36.2 & 65.9 & 23.4 & 70.4 & 52.2 & 70.9 & 59.2 \\
			SENet+~\cite{hu2018squeeze} & \checkmark &  & \checkmark & 38.9 & 78.9 & 36.9 & 65.4 & 22.9 & 71.9 & 53.5 & 71.8 & 61.3 \\
			Ellipse R-CNN* & \checkmark & \checkmark & \checkmark & 40.0 & 81.4 & 37.8 & 64.2 & 21.2 & \textbf{68.0} & 54.2 & 73.7 & 63.7 \\
			Ellipse R-CNN & \checkmark & \checkmark & \checkmark & \textbf{41.2} & \textbf{83.9} & \textbf{39.0} & \textbf{63.4} & \textbf{19.0} & 68.5 & \textbf{56.4} & \textbf{76.5} & \textbf{66.8} \\
			\bottomrule
		\end{tabular}
	\end{center}
\end{table*}

\section{Experiments} \label{sec:experiments}
In this section, we first introduce synthetic and real datasets we use for the experiments, followed by a description of the implementation details and evaluation metrics.
After that, we show experimental results of the ablation study for our Ellipse R-CNN detector, and make a comparison to the state of the art.
In the end, we demonstrate how Ellipse R-CNN helps improve the accuracy of 3D object estimation in occluded cases.

\subsection{Datasets}
We validate the proposed Ellipse R-CNN on four datasets: synthetic occluded ellipses (SOE), synthetic occluded fruits (SOF), real occluded fruits (ROF) and FDDB~\cite{jain2010fddb} datasets.
Each elliptical object is annotated by its five ellipse parameters of the whole object region along with a bounding box of the visible part (except for the FDDB dataset as shown in Fig.~\ref{fig:ellipseRegression}).

The SOE dataset consists of 16,500 images in total, approximately 15,000 images are for training and the rest for testing.
Synthetic images are generated from a cluster of different ellipses occluded from each other in the same distribution as in Fig.~\ref{fig:ellipseAngleCompare}.
The image background is randomly filled by the Pascal dataset~\cite{everingham2010pascal} with randomly added triangles (simulating nearby obstacles) to further occlude ellipses (the visibility ratio of each $r_o \in [0.3, 0.6]$).
To introduce more interference, ellipse colors are randomly generated in a roughly same tone as in real cases (e.g., clustered fruits and faces).

The SOF dataset contains 3,545 images (3,040 for training and 505 for testing) of a cluster of fruits occluded in a realistic tree ($r_o \ge 0.3$), which are generated by changing different poses and sizes of each model in Unreal Engine (UE) with the background randomly filled by images taken from different real orchards~\cite{dong2018semantic}.
The GT ellipses are obtained by projecting the 3D fruit ellipsoids onto the corresponding images~\cite{qiu2017unrealcv} based on camera poses.

The ROF dataset (1115 images in total) is human-annotated and is built upon MinneApple~\cite{hani2019minneapple} and ACFR~\cite{bargoti2017deep} datasets, from which we crop out the sub-images of heavily occluded fruit clusters. We perform a similar training-and-test split as in~\cite{bargoti2017deep}, which are composed of 900 images and 215 images, respectively.
FDDB dataset~\cite{jain2010fddb} includes 2,845 images of 5,171 faces that are split by ten folds.
Since most faces are well-separated and only have GT ellipses (without GT visible boxes), we just demonstrate the generalization of our ellipse regressor on this dataset through 10-fold cross-validation.

\begin{figure*}[!t]
	\centering
	\includegraphics[width=0.99\textwidth]{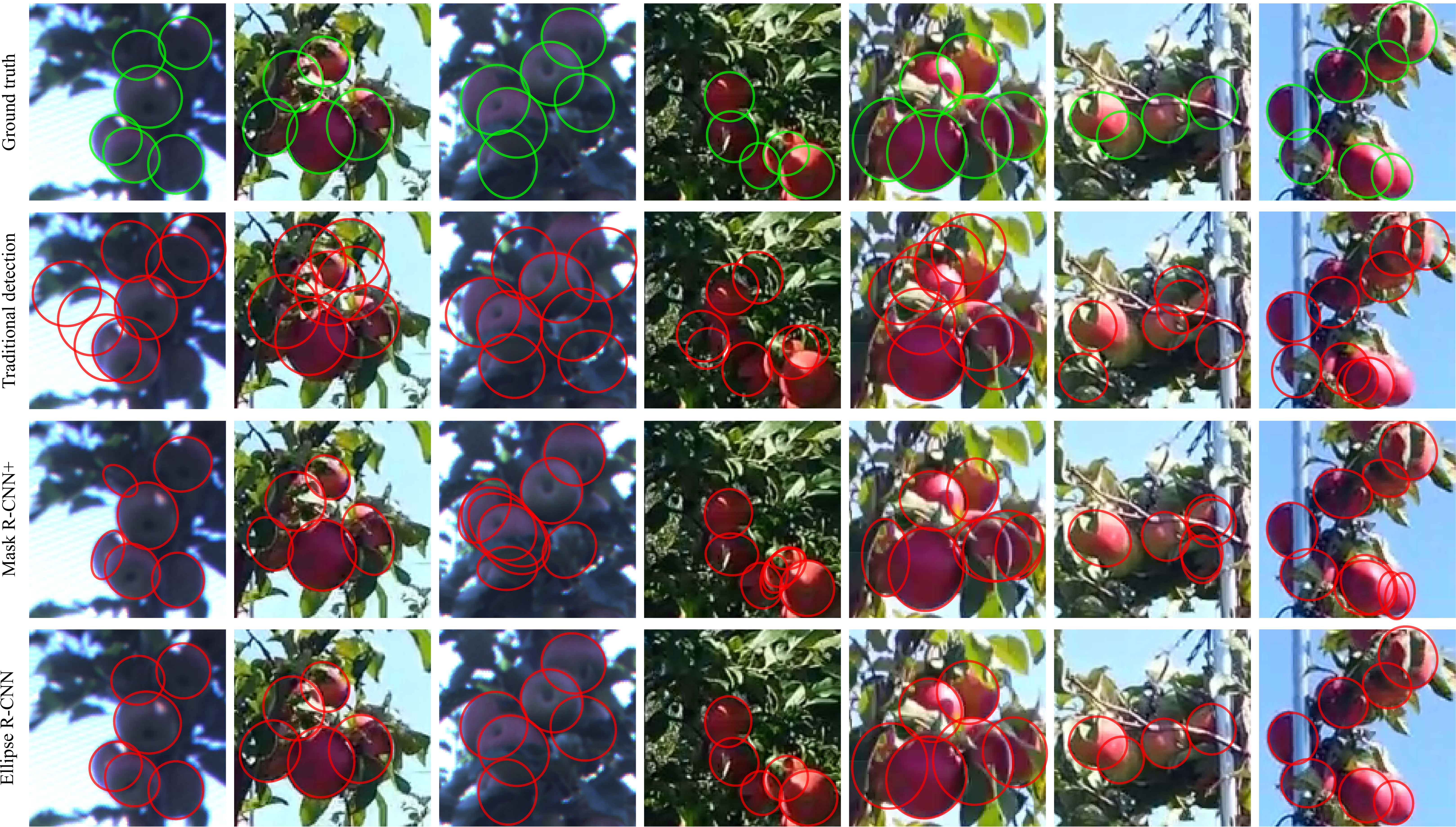}
	\caption{Qualitative results from our Ellipse R-CNN, Mask R-CNN+, and traditional ellipse detection~\cite{xie2002new} on the ROF dataset. The GT ellipses are human-annotated. Since most fruit shapes are close to circles, the orientation information of annotated ellipses is noisy. Here, we mainly focus on the performance evaluated by AP and MR as shown in Table~\ref{tab:real}. The Mask R-CNN+ trained on the whole object regions outputs many false positives in heavily occluded cases. The traditional detection method even with prior information of fruit shapes fails to detect ellipsoid objects from the image.}
	\label{fig:expResReal}
\end{figure*}
\begin{table*}
	\caption{Overall performance on the ROF dataset. $\text{AP}_{\star}$ and $\text{MR}_{\star}$: the ellipse IoU level starts from 0.7 to 0.9 with an interval 0.05. \newline $\text{AP}_{\star}^{\Theta}$: the angle error decreases from $45^{\circ}$ to $5^{\circ}$ with an interval $10^{\circ}$. The default ellipse IoU for $\text{AP}^{\Theta}$ is 0.7.} \label{tab:real}
	\begin{center}
		\begin{tabular}{l|M{.22cm} M{.22cm} M{.22cm}|M{.7cm} M{.7cm} M{.7cm}|M{.7cm} M{.7cm} M{.7cm}|M{.7cm} M{.7cm} M{.7cm}}
			\toprule
			Methods & \textbf{R} & \textbf{O} & \textbf{A} & $\text{AP}_{\star}$ & $\text{AP}_{70}$ & $\text{AP}_{80}$ & $\text{MR}_{\star}$ & $\text{MR}_{70}$ & $\text{MR}_{80}$ & $\text{AP}_{\star}^{\Theta}$ & $\text{AP}_{45}^{\Theta}$ & $\text{AP}_{25}^{\Theta}$ \\
			\bottomrule\toprule
			Mask R-CNN+ & -- & -- & -- & 33.4 & 59.8 & 34.2 & 69.6 & 44.1 & 69.2 & 20.0 & 34.8 & 20.6 \\
			Ellipse R-CNN- &  &  &  & 36.1 & 64.1 & 37.1 & 67.0 & 39.9 & 66.4 & 24.5 & 40.5 & 25.9 \\
			Ellipse R-CNN- & \checkmark &  &  & 40.5 & 68.6 & 42.0 & 63.5 & 34.7 & 61.6 & 31.0 & 49.7 & 33.2 \\
			\midrule
			DeepParts+~\cite{tian2015deep} & \checkmark & \checkmark &  & 43.2 & 74.5 & 44.4 & 61.4 & 29.2 & 59.2 & 34.8 & 53.6 & 37.6 \\
			SENet+~\cite{hu2018squeeze} & \checkmark &  & \checkmark & 43.9 & 74.0 & 46.5 & 60.9 & 28.5 & 59.7 & 36.9 & 55.1 & 39.7 \\
			Ellipse R-CNN* & \checkmark & \checkmark & \checkmark & 44.6 & 76.4 & \textbf{49.2} & 59.1 & 26.3 & \textbf{56.2} & 38.6 & 57.8 & 42.8 \\
			Ellipse R-CNN & \checkmark & \checkmark & \checkmark & \textbf{45.8} & \textbf{78.2} & 48.9 & \textbf{58.1} & \textbf{24.6} & 56.9 & \underline{40.7} & \underline{60.0} & \underline{44.2} \\
			\bottomrule
		\end{tabular}
	\end{center}
\end{table*}

\subsection{Implementation Details}
We use TensorFlow~\cite{abadi2016tensorflow} to implement and train the Ellipse R-CNN.
For comparison, we directly use the source code of Mask R-CNN provided by Matterport~\cite{matterport_maskrcnn_2017}.
For the training, we use the pre-trained weights for MS COCO~\cite{lin2014microsoft} to initialize the Ellipse R-CNN, and use a step strategy with mini-batch stochastic gradient descent (SGD) to train the networks on a GeForce GTX 1080 GPU.
On SOF, ROF, and FDDB datasets, we train with an initial learning rate of $10^{-3}$ for 20,000 iterations and train for another 10,000 iterations with a decreased learning rate of $10^{-4}$.
On the SOE dataset, we start with the same learning rate of $10^{-3}$, and then decrease the learning rate by 5 after every 20,000 iterations.
The model converges at 50,000 iterations.
During the training, we perform on-the-fly data augmentation with flipping, shifting, and rotation at random.
We resize the ellipse and fruit images to 128$\times$128, while the face images are resized to 256$\times$256 in order to have face details in a higher resolution for training and testing.

\begin{figure*}[!t]
	\centering
	\includegraphics[width=0.99\textwidth]{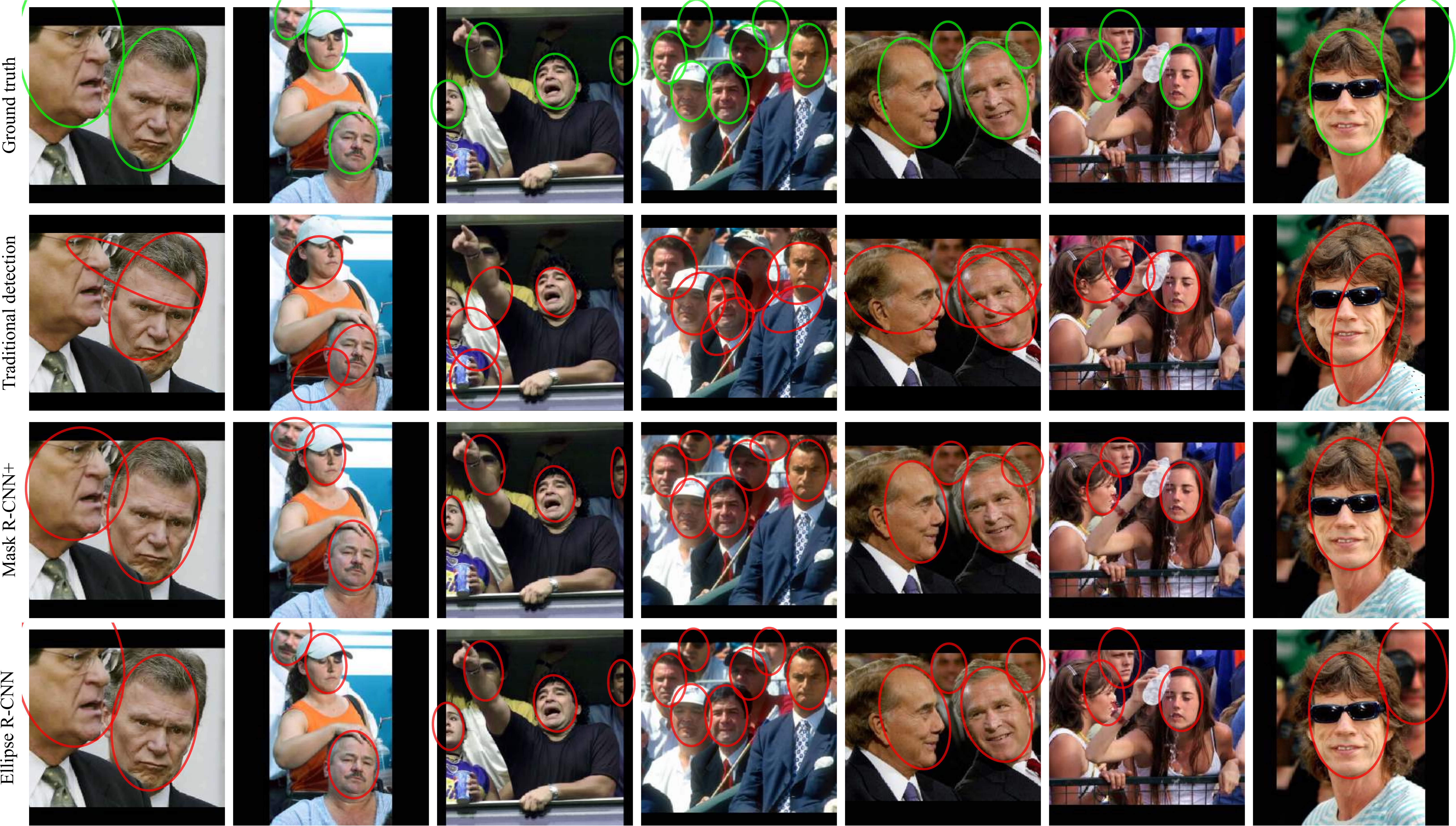}
	\caption{Qualitative results from our Ellipse R-CNN, Mask R-CNN+, and traditional ellipse detection~\cite{xie2002new} on the FDDB~\cite{jain2010fddb} dataset. Compared to the Mask R-CNN+, the Ellipse R-CNN is able to infer the whole face regions even they are occluded around the image boundaries. The traditional detection method even with prior information of face shapes fails to detect ellipsoid objects from the image.}
	\label{fig:expResFDDB}
\end{figure*}
\begin{table*}
	\caption{10-fold cross-validation on the FDDB dataset. $\star$: the angle error decreases from $10^{\circ}$ to $2^{\circ}$ with an interval $2^{\circ}$. \newline The default ellipse IoU for $\text{AP}^{\Theta}$ and $\text{MR}^{\Theta}$ is 0.75.} \label{tab:fddb}
	\begin{center}
		\begin{tabular}{l|c|M{.22cm}|M{.54cm} M{.54cm} M{.54cm} M{.54cm} M{.54cm} M{.54cm} M{.54cm} M{.54cm} M{.54cm} M{.54cm}|M{.54cm}}
			\toprule
			Methods & Metrics & \textbf{R} & F-1 & F-2 & F-3 & F-4 & F-5 & F-6 & F-7 & F-8 & F-9 & F-10 & Avg. \\
			\bottomrule\toprule
			Mask R-CNN+ &  & -- & 59.0 & 56.9 & 60.0 & 61.4 & 56.7 & 58.2 & 59.3 & 59.3 & 55.1 & 61.1 & 58.7 \\
			Ellipse R-CNN- & $\text{AP}_{\star}^{\Theta}$ &  & 64.4 & 65.1 & 66.3 & 67.5 & 66.8 & 67.6 & 67.1 & 66.5 & 65.2 & 66.9 & 66.3 \\
			Ellipse R-CNN- &  & \checkmark & \textbf{68.7} & \textbf{70.9} & \textbf{71.4} & \textbf{72.2} & \textbf{72.7} & \textbf{74.0} & \textbf{73.0} & \textbf{71.8} & \textbf{71.9} & \textbf{71.1} & \textbf{71.8} \\
			\midrule
			Mask R-CNN+ &  & -- & 21.2 & 21.0 & 18.5 & 18.6 & 17.6 & 14.8 & 20.0 & 18.2 & 22.2 & 18.8 & 19.1 \\
			Ellipse R-CNN- & $\text{MR}_{\star}^{\Theta}$ &  & 16.8 & 15.9 & 15.2 & 14.5 & 14.3 & 13.7 & 14.9 & 14.5 & 15.7 & 15.2 & 15.1 \\
			Ellipse R-CNN- &  & \checkmark & \textbf{11.2} & \textbf{11.9} & \textbf{11.1} & \textbf{10.2} & \textbf{10.6} & \textbf{8.4} & \textbf{9.6} & \textbf{9.8} & \textbf{11.5} & \textbf{10.6} & \textbf{10.5} \\
			\bottomrule
		\end{tabular}
	\end{center}
\end{table*}

\subsection{Evaluation Metrics}
Four evaluation metrics are exploited in all of our experiments: average precision (AP~\cite{lin2014microsoft} over ellipse IoU thresholds), log-average miss rate (MR)~\cite{dollar2011pedestrian}, $\text{AP}^{\Theta}$ and $\text{MR}^{\Theta}$ (AP and MR over ellipse angle errors).
MR is the average value of miss rates for 9 FPPI (false positives per image) rates evenly spaced in the log-space ranging from $10^{-2}$ to $10^{0}$.
By introducing $\text{AP}^{\Theta}$ and $\text{MR}^{\Theta}$, we focus more on the accuracy of predicted ellipse angles.
For example, we consider a prediction (evaluated by $\text{AP}_{45}^{\Theta}$ or $\text{MR}_{45}^{\Theta}$) as a false positive if its ellipse IoU is less than 0.75 (set as the default IoU) or its angle error is greater than $45^{\circ}$.
To clearly show the performance difference, we use strict criteria: for instance, the IoU level starts from 0.75 up to 0.95 with an interval 0.05 (e.g., $\text{AP}_{75:95}$ written as $\text{AP}_{\star}$), and the angle error decreases from $45^{\circ}$ to $5^{\circ}$ with an interval $10^{\circ}$ (e.g., $\text{MR}_{45:5}^{\Theta}$ written as $\text{MR}_{\star}^{\Theta}$).
We use AP and MR to measure the overall performance as they place a significantly large emphasis on localization and miss rate in occluded cases, respectively.

\subsection{Performance of Ellipse R-CNN}
We compare the proposed Ellipse R-CNN to the baseline model Mask R-CNN, which obtains the state-of-the-art results in general object detection and instance segmentation.
Since our model is the first work of ellipse regression, to make a fair comparison, we fit ellipses directly from the mask outputs of Mask R-CNN (trained on the regions of whole objects) using the method of minimum volume enclosing ellipsoid~\cite{moshtagh2005minimum} in 2D (i.e., Mask R-CNN+).
We run a number of ablations to further analyze Ellipse R-CNN.
For the ablation study of occlusion handling, we adapt two state-of-the-art methods in our model: DeepParts+ and SENet+.
In DeepParts+, we only keep the U-Net structure to learn a set of 45 occlusion patterns, and the final score is obtained via an MLP on the part detection scores~\cite{tian2015deep}.
For SENet+, we learn the attention vector directly from the refined feature maps (without U-Net), and perform the classification only on the re-weighted features~\cite{hu2018squeeze}.
The same strategy for non-maximum suppression in the previous work~\cite{ren2017faster, hemask} is applied to the predicted regions of each model.

\subsubsection{Accuracy of Ellipse Regression}
\begin{figure*}[!t]
	\centering
	\includegraphics[width=0.99\textwidth]{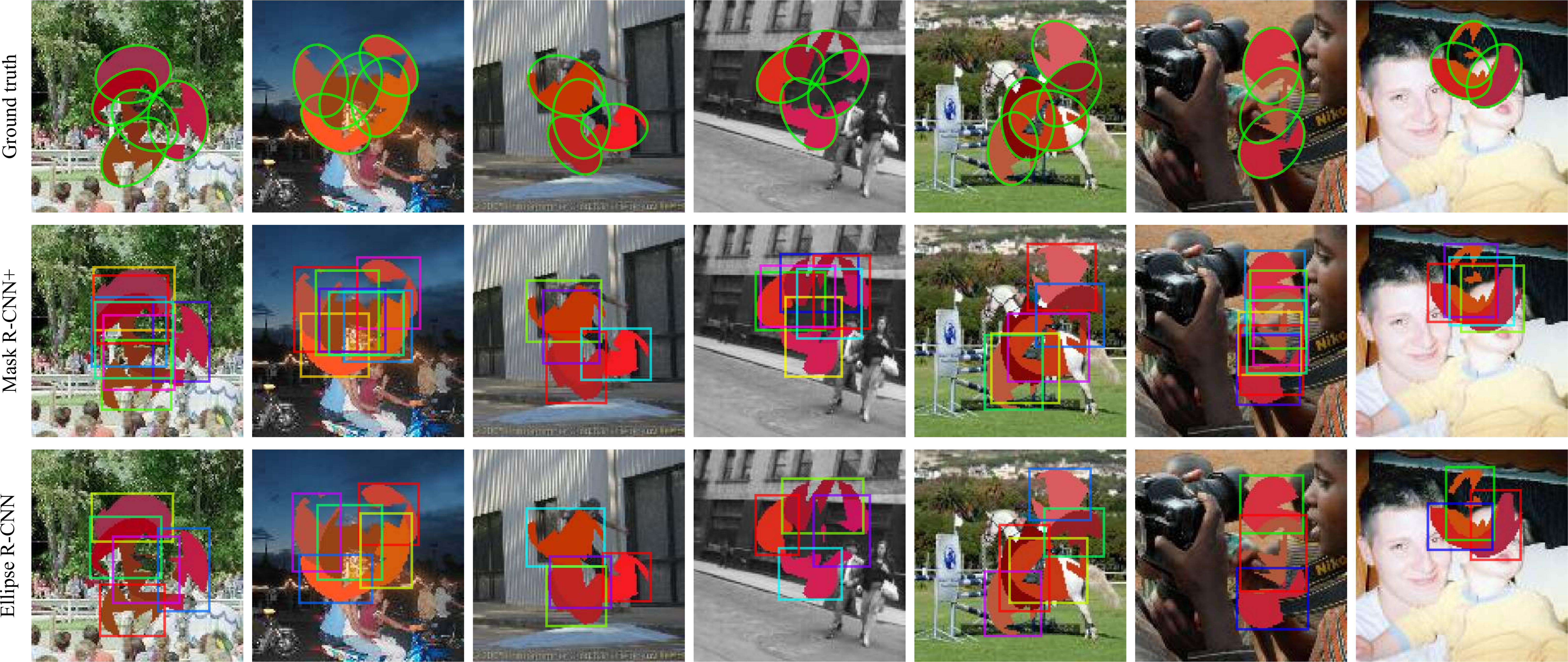}
	\caption{Qualitative results of predicted object regions from our Ellipse R-CNN (trained on visible regions) and Mask R-CNN+ (trained on entire object regions) on the SOE dataset. Ellipse R-CNN outputs much smaller regions with fewer overlaps, especially for occluded objects, while Mask R-CNN+ suffers from outputting more false positives due to high similarities among the proposed entire object regions. Predicted regions are colored randomly for better visualization from large overlaps.}
	\label{fig:expBboxEllipse}
\end{figure*}

\begin{figure*}[!t]
	\centering
	\includegraphics[width=0.99\textwidth]{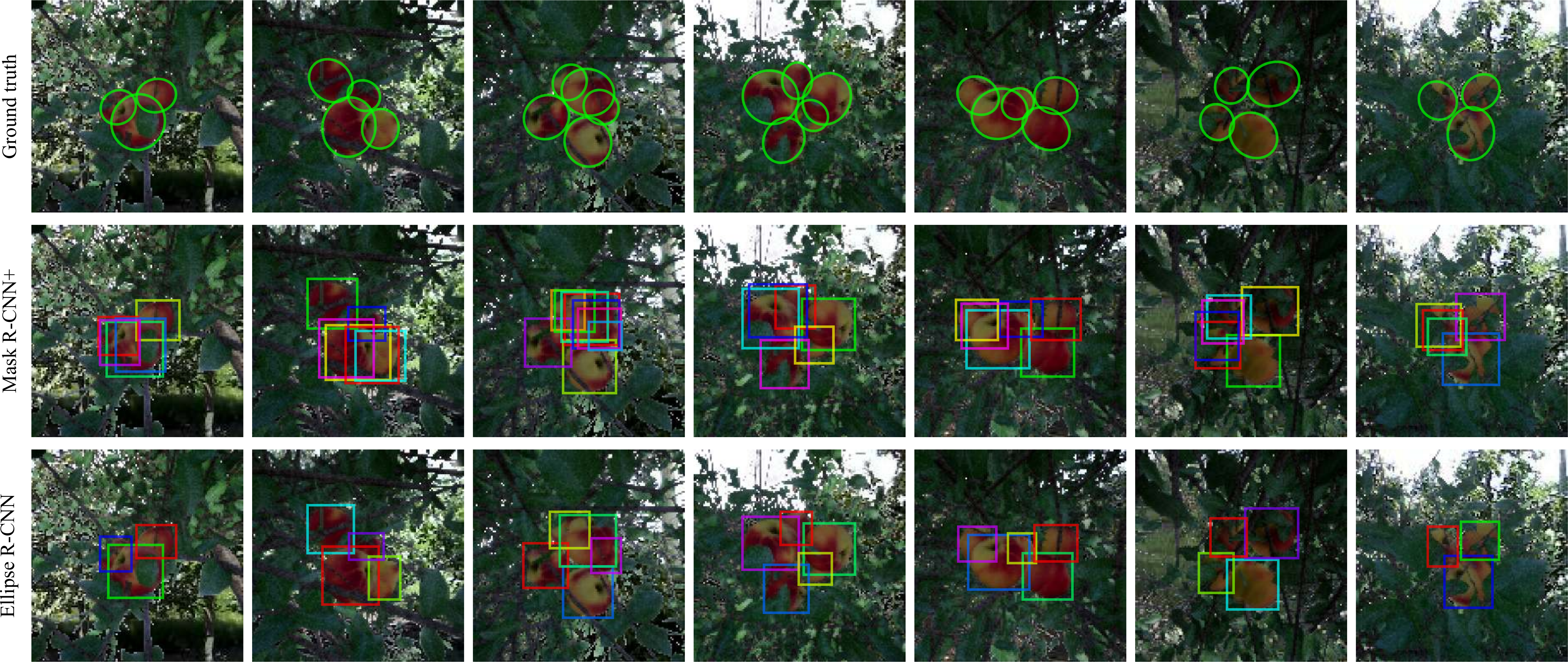}
	\caption{Qualitative results of predicted object regions from our Ellipse R-CNN (trained on visible regions) and Mask R-CNN+ (trained on entire object regions) on the SOF dataset. Compared to Mask R-CNN+, Ellipse R-CNN focuses on smaller regions (visible parts) with fewer overlaps for occluded objects.}
	\label{fig:expBboxUE}
\end{figure*}

\begin{figure*}[!t]
	\centering
	\includegraphics[width=0.99\textwidth]{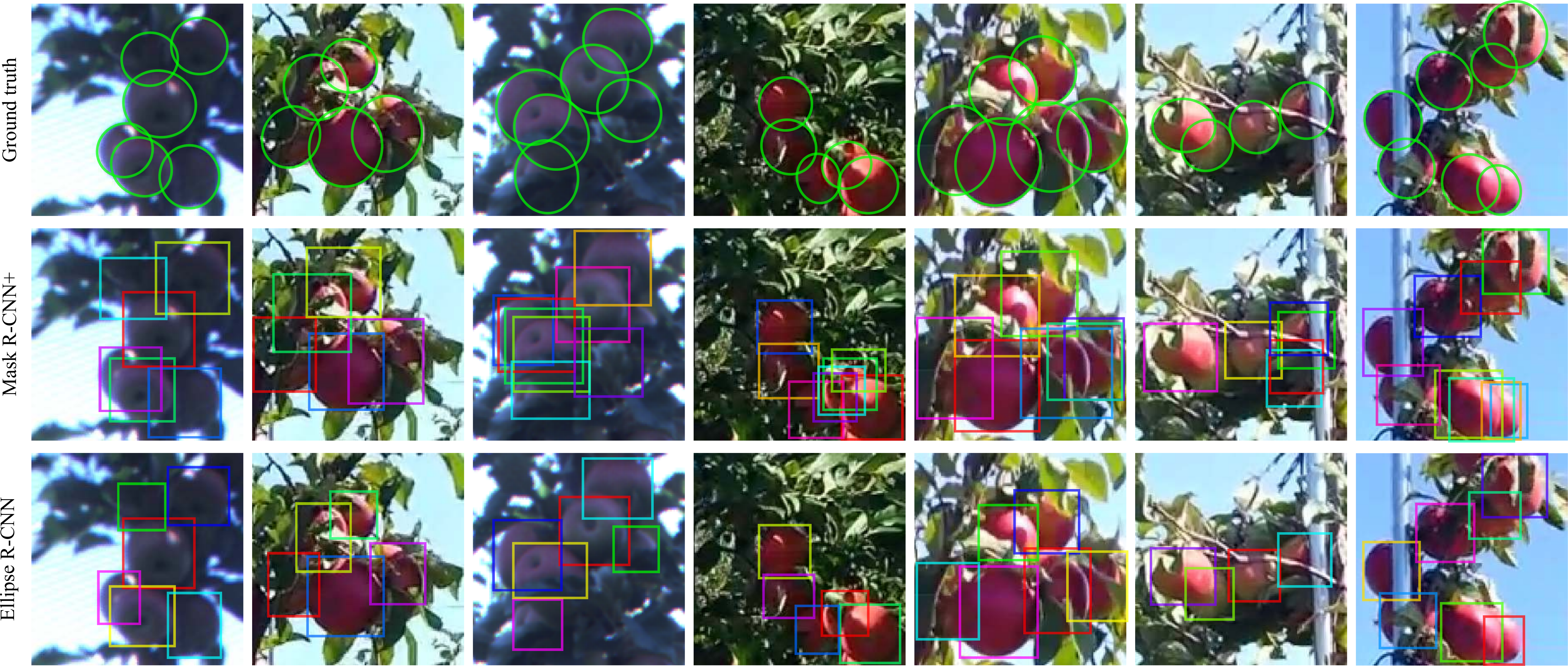}
	\caption{Qualitative results of predicted object regions from our Ellipse R-CNN (trained on visible regions) and Mask R-CNN+ (trained on entire object regions) on the ROF dataset. Overall, our model predicts smaller object regions with fewer overlaps in occluded scenarios, which verifies the effectiveness of Ellipse R-CNN for inferring entire shapes of occluded objects directly from the visible parts.}
	\label{fig:expBboxReal}
\end{figure*}

The key component of our Ellipse R-CNN is the ellipse regressor.
Some examples of detected elliptical objects are illustrated in Fig.~\ref{fig:expResEllipse}--\ref{fig:expResFDDB}.
Table~\ref{tab:ellipse}--\ref{tab:ue} show the breakdown performance of the ellipse prediction on the SOE and SOF datasets whose GT is perfectly generated based on the geometry of object models.
Our strategy of ellipse regression (e.g., ellipse R-CNN-) leads significant performance improvement on all metrics compared to the baseline model.
Specifically, Table~\ref{tab:ellipse} shows that both $\text{AP}^{\Theta}$ and $\text{MR}^{\Theta}$ values of the proposed model are not sensitive to the increased levels of angle errors, which means that our strategy achieves a high accuracy of ellipse orientation prediction.
We also observe that the Mask R-CNN+ model trained on whole object regions (instead of visible parts) suffers from outputting more false positives due to the high similarities among the proposed feature regions (see Fig.~\ref{fig:expResEllipse}--\ref{fig:expResReal}).
For the ROF dataset, Table~\ref{tab:real} shows a higher sensitivity of our model on $\text{AP}^{\Theta}$ and $\text{MR}^{\Theta}$ compared to those on SOE and SOF datasets: $\text{AP}_{\star}$ is higher than that in Table~\ref{tab:ue} but $\text{AP}_{\star}^{\Theta}$ drops a lot. The reason is that most human-annotated fruits are close to circles whose GT orientation information is noisy and inconsistent. Thus, it is hard to quantify the results on $\text{AP}^{\Theta}$ and $\text{MR}^{\Theta}$ but our proposed model still achieves the best performance on AP and MR.
Fig.~\ref{fig:expBboxEllipse}--\ref{fig:expBboxReal} further demonstrate the predicted object regions from both Mask R-CNN+ and Ellipse R-CNN.
We observe that the predicted regions from our Ellipse R-CNN are overall much smaller with fewer overlaps than those from Mask R-CNN+, especially for occluded objects, which verifies the effectiveness of Ellipse R-CNN for inferring entire shapes of occluded objects directly from their visible parts.

\begin{figure}[!t]
	\centering
	\includegraphics[width=0.99\columnwidth]{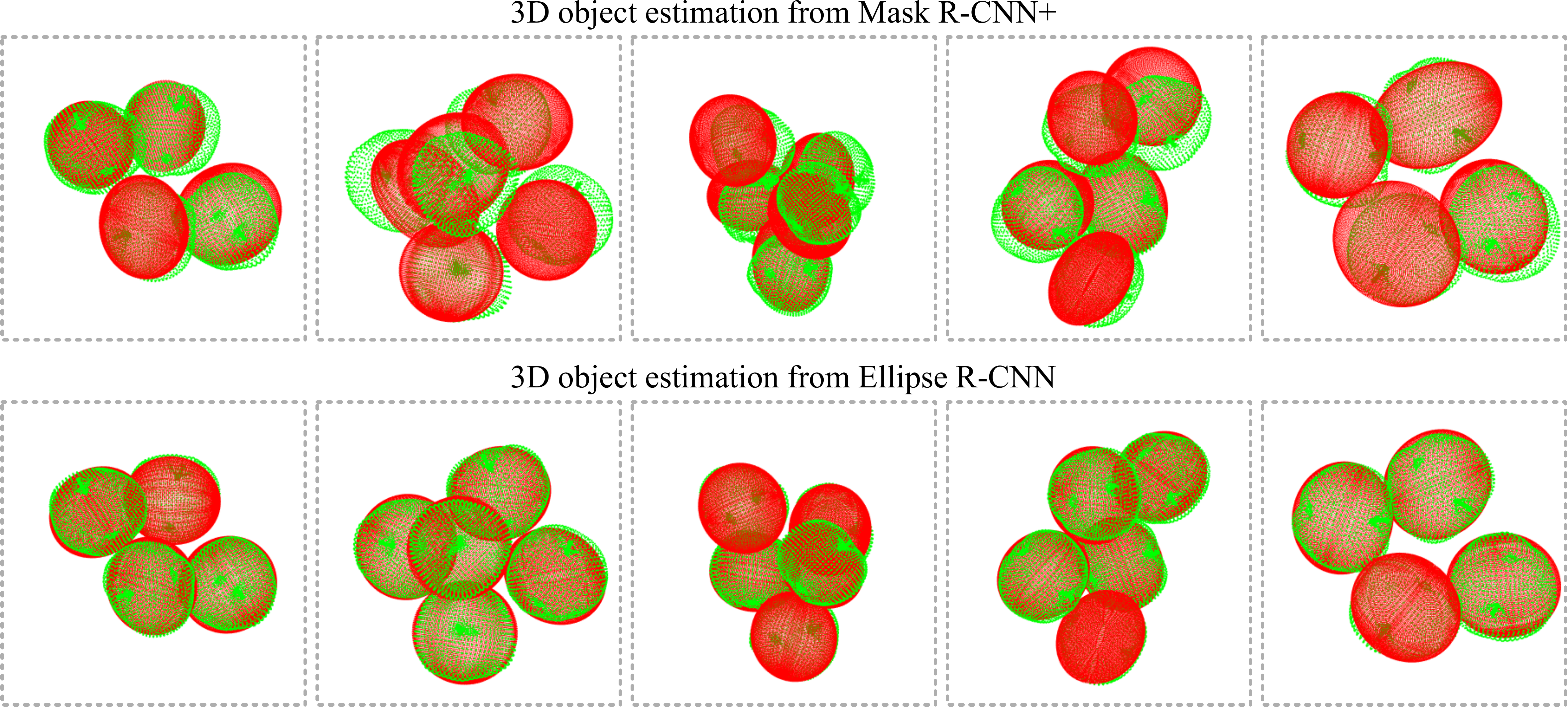}
	\caption{Qualitative results of 3D object estimation from our Ellipse R-CNN and Mask R-CNN+ on the SOF dataset. The five different settings correspond to the columns from the 2nd to the 6th in Fig.~\ref{fig:expResUE}. False positives are removed for Mask R-CNN+. The estimation results are red-colored while the 3D GT points are in green.}
	\label{fig:expRes3d}
\end{figure}

\begin{table}[t!]
	\caption{3D estimation errors on the SOF dataset. The relative size error is the average over each dimension compared to the GT.} \label{tab:3D}
	\begin{center}
		\begin{tabular}{l|M{1.78cm}|M{1.78cm}|M{1.78cm}}
			\toprule
			Methods & Rot. Error ($^{\circ}$) & Pos. Error (cm) & Rel. Size Error \\
			\bottomrule\toprule
			Mask R-CNN+ & 37.2 & 3.3 & 28.5\% \\
			Ellipse R-CNN & 12.6 & 1.6 & 10.3\% \\
			\bottomrule
		\end{tabular}
	\end{center}
\end{table}

\subsubsection{Validity of Feature Region Refinement}
Table~\ref{tab:ellipse}--\ref{tab:real} show the detailed breakdown performance of the proposed feature region refinement (i.e., Ellipse R-CNN- with \textbf{R}) on the SOE, SOF and ROF datasets.
The performance is largely improved when the refined features are used for ellipse regression and classification.
The improvements in $\text{AP}^{\Theta}$ and $\text{MR}^{\Theta}$ indicate that the refinement strategy is not only beneficial to increasing the accuracy of ellipse region prediction but also to reducing the false positives for classification, especially in occluded cases.
However, Table~\ref{tab:fddb} shows smaller improvements if we apply the feature refinement strategy on the FDDB dataset: $\text{AP}_{\star}^{\Theta}$ and $\text{MR}_{\star}^{\Theta}$ are only improved by 5.5 and 4.6, respectively.
As discussed in Sec.~\ref{subsec:featureRefinement}, feature region refinement is used to remove the interference of nearby occlusions.
Most faces in the FDDB datasets are well-separated and there are few clustered and occluded cases.
Thus, the improvements in Table~\ref{tab:fddb} by using the refined features are not as significant as those in Table~\ref{tab:ellipse}--\ref{tab:real}.

\subsubsection{Performance of Occlusion Handling}
One of our evaluation goals is occlusion handling, whose overall performance is measured by $\text{MR}$ and $\text{MR}^{\Theta}$ as shown in Table~\ref{tab:ellipse}--\ref{tab:real}.
All three variants with different mechanisms of occlusion handling show some improvements to the baseline (i.e., Ellipse R-CNN- with \textbf{R}), ranging from 2.1 to 7.2 on $\text{MR}_{\star}$ and from 4.3 to 8.6 on $\text{MR}_{\star}^{\Theta}$.
Overall, the error rates can be sorted in the following order: DeepParts+ $\approx$ SENet+ $>$ Ellipse R-CNN*.
The reason is that the DeepParts+ is limited by its fixed number of occlusion patterns to learn, while the SENet+ learns a continuous attention vector to adjust feature weights but lacks the whole ellipse information to generalize different occlusions.
We further compare our Ellipse R-CNN to the Ellipse R-CNN* (without concatenating $f_e$).
The gap between them demonstrates that our concatenation of $f_c$ with $f_e$ is a more effective way of generalizing various occlusion patterns from ellipse predictions.

\subsubsection{Generalization of Ellipse Regressor}
In order to investigate the generalization ability of the proposed ellipse regressor, we also perform experiments on the FDDB dataset.
Since no GT visible boxes are available and few objects are clustered, we can only evaluate our model without the occlusion handling mechanism (i.e., Ellipse R-CNN- with \textbf{R}).
Focusing on the accuracy of orientation prediction, we show the results of 10-fold cross-validation in Table~\ref{tab:fddb}, where we can see that our model outperforms the Mask R-CNN+ baseline by 13.1 on $\text{AP}_{\star}^{\Theta}$ and 8.6 on $\text{MR}_{\star}^{\Theta}$, respectively.
We also show some qualitative results in Fig.~\ref{fig:expResFDDB}, where we can observe that our detector produces robust detections of ellipses even in some extreme cases.
Specifically, in all seven examples, several faces are heavily occluded by the image boundaries. The Mask R-CNN+ produces many distorted face shapes, while our detector accurately infers the whole ellipse regions for all of them.

\subsubsection{Discussion on 3D Object Estimation}
In order to understand how Ellipse R-CNN improves the accuracy of 3D object estimation, we implement the multi-view 3D localization using quadrics~\cite{rubino20183d} from 2D detections on the SOF dataset.
We compare our detector with the Mask R-CNN and summarize the results in Table~\ref{tab:3D}.
The evaluation metrics include rotation error~\cite{dong2018novel}, position error and relative size error in 3D that are averaged over all objects.
For each UE setting (24 different settings in total), we select three images taken from different view angles to serve as the same inputs for both methods.
As shown in the comparison, three estimation errors of the Ellipse R-CNN are much lower than the Mask R-CNN+, especially the rotation error (i.e., $12.6^{\circ}$ vs. $37.2^{\circ}$).
This is because Ellipse R-CNN better infers the whole region of each object directly from the visible part, thus is more effective in estimating the 3D pose and shape of objects from occlusion.
More qualitative results are shown in Fig.~\ref{fig:expRes3d}.

%% file: tipAppendices.tex
\section*{Appendices} \label{sec:appendices}

\subsection{Qualitative Results in Ablation Studies}
\begin{figure*}[!t]
	\centering
	\includegraphics[width=0.99\textwidth]{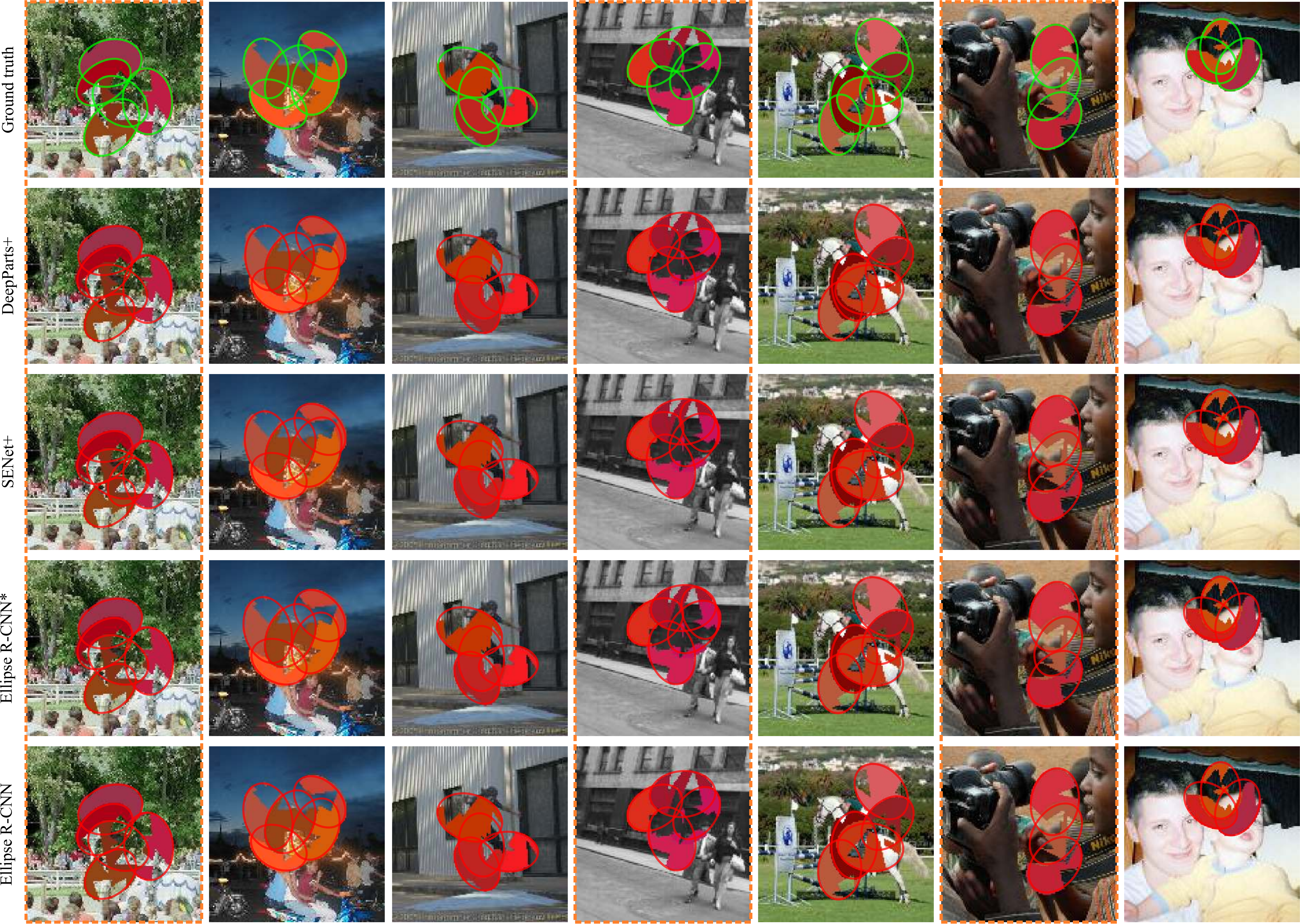}
	\caption{Qualitative results from our Ellipse R-CNN, Ellipse R-CNN*, SENet+, and DeepParts+ on the SOE dataset. Column 1, Column 4, and Column 6 are highlighted in yellow dashed boxes for clear visualization of their performance differences.}
	\label{fig:expAblationEllipse}
\end{figure*}

\begin{figure*}[!t]
	\centering
	\includegraphics[width=0.99\textwidth]{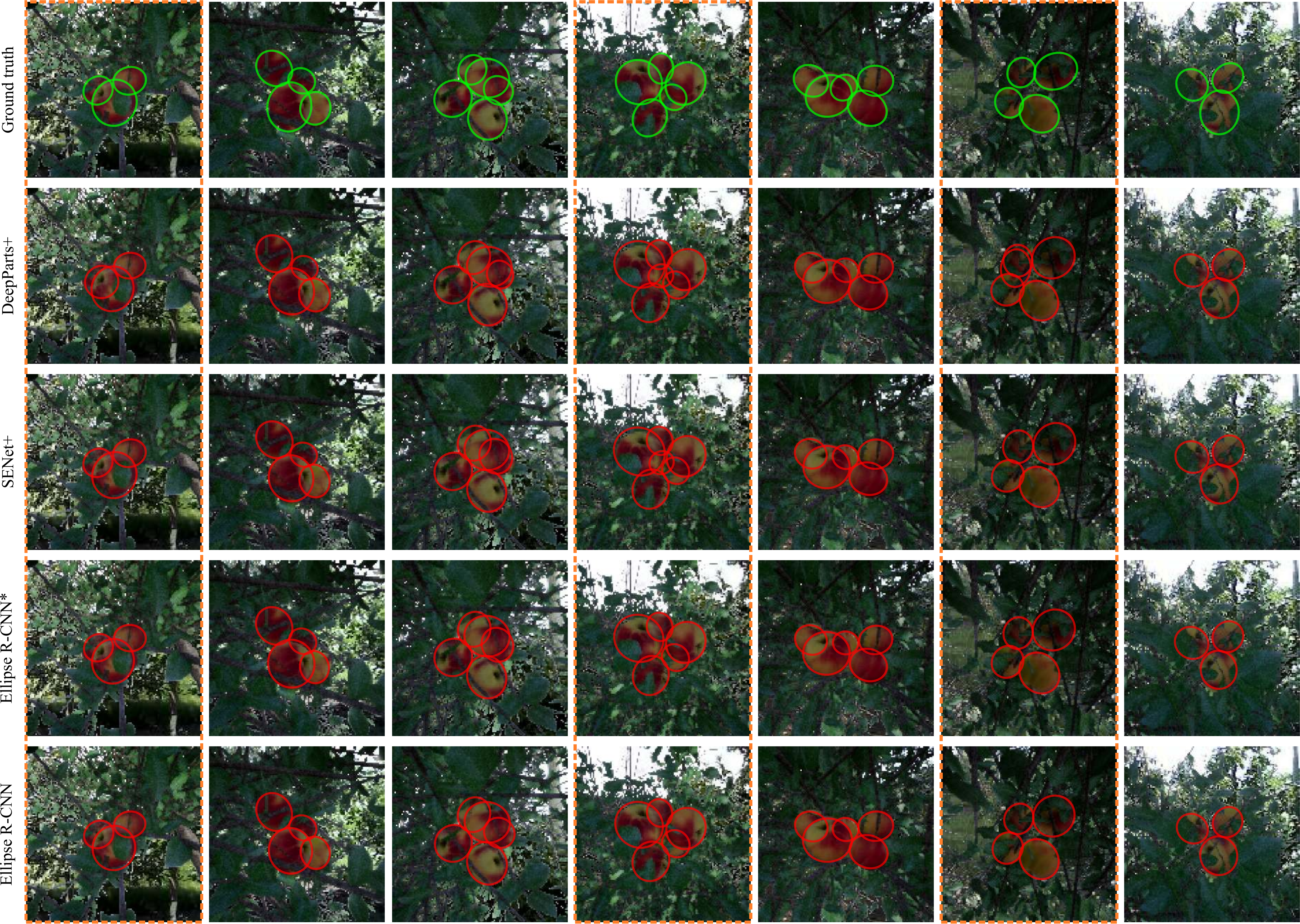}
	\caption{Qualitative results from our Ellipse R-CNN, Ellipse R-CNN*, SENet+, and DeepParts+ on the SOF dataset. Column 1, Column 4, and Column 6 are highlighted in yellow dashed boxes for clear visualization of their performance differences.}
	\label{fig:expAblationUE}
\end{figure*}

\begin{figure*}[!t]
	\centering
	\includegraphics[width=0.99\textwidth]{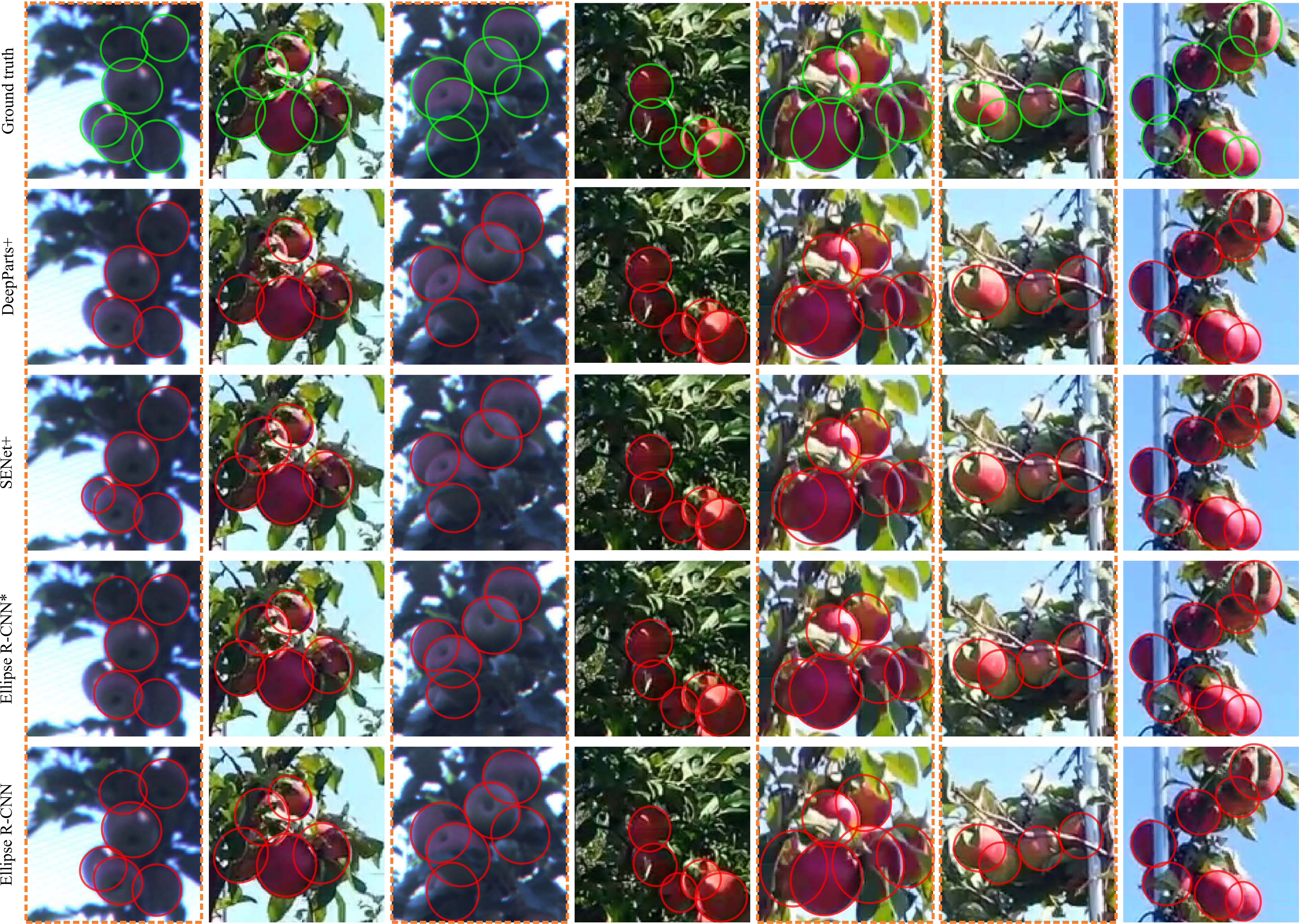}
	\caption{Qualitative results from our Ellipse R-CNN, Ellipse R-CNN*, SENet+, and DeepParts+ on the ROF dataset. Column 1, Column 3, Column 5, and Column 6 are highlighted in yellow dashed boxes for clear visualization of their performance differences.}
	\label{fig:expAblationReal}
\end{figure*}

Fig.~\ref{fig:expAblationEllipse}--\ref{fig:expAblationReal} demonstrate qualitative results from our Ellipse R-CNN, Ellipse R-CNN*, SENet+~\cite{hu2018squeeze}, and DeepParts+~\cite{tian2015deep} on SOE, SOF, and ROF datasets, respectively.
We highlight some examples using yellow dashed boxes for readers to clearly visualize the performance difference among such methods in the ablation studies.
We observe that DeepParts+, SENet+, and Ellipse R-CNN* all outperform Mask R-CNN+ by using different strategies for handling occluded scenarios.
Overall, the performances can be sorted in the following order: Mask R-CNN+ $<$ DeepParts+ $\approx$ SENet+ $<$ Ellipse R-CNN* $<$ Ellipse R-CNN.
According to Table~\ref{tab:ellipse}--\ref{tab:real} and Fig.~\ref{fig:expAblationEllipse}--\ref{fig:expAblationReal}, our Ellipse R-CNN achieves the best performance in terms of both the accuracy of regressed ellipses and the effectiveness of reducing false positives.

\begin{table*}
	\caption{10-fold cross-validation on the FDDB dataset using $\text{AP}_{50:5:95}$. $\textbf{R}$: ellipse regression with feature refinement. \newline $\text{AP}_{\star}$: the pixel IoU level of ellipse mask starts from 0.5 to 0.95 with an interval 0.05.} \label{tab:fddbStandard}
	\begin{center}
		\begin{tabular}{l|l|M{.22cm}|M{.54cm} M{.54cm} M{.54cm} M{.54cm} M{.54cm} M{.54cm} M{.54cm} M{.54cm} M{.54cm} M{.54cm}|M{.54cm}}
			\toprule
			Methods & Metrics & \textbf{R} & F-1 & F-2 & F-3 & F-4 & F-5 & F-6 & F-7 & F-8 & F-9 & F-10 & Avg. \\
			\bottomrule\toprule
			& $\text{AP}_{50}$ &  & 91.8 & 92.0 & 91.9 & 91.9 & 91.5 & 92.4 & 91.7 & 92.1 & 90.9 & 92.5 &  92.0 \\
			Mask R-CNN & $\text{AP}_{75}$ & -- & 79.1 & 81.3 & 82.5 & 83.2 & 83.1 & 85.2 & 82.2 & 82.9 & 78.9 & 82.9 & 82.1 \\
			& $\text{AP}_{\star}$ &  & 64.8 & 64.7 & 65.2 & 64.7 & 65.2 & 65.6 & 65.0 & 65.5 & 63.7 & 65.6 & 65.0 \\
			\midrule
			& $\text{AP}_{50}$ &  & 92.8 & 92.9 & 93.4 & 93.1 & 93.2 & 93.4 & 92.9 & 93.3 & 92.9 & 93.7 & \textbf{93.2} \\
			Ellipse R-CNN- & $\text{AP}_{75}$ & \checkmark & 88.7 & 88.3 & 89.2 & 89.6 & 89.9 & 91.1 & 90.2 & 90.5 & 89.0 & 90.4 & \textbf{89.7} \\
			& $\text{AP}_{\star}$ &  & 79.1 & 78.9 & 79.8 & 79.1 & 79.7 & 80.5 & 79.8 & 80.1 & 78.2 & 79.5 & \textbf{79.5} \\
			\bottomrule
		\end{tabular}
	\end{center}
\end{table*}

\subsection{Training Mask R-CNN on Visible Parts}
\begin{figure*}[!t]
	\centering
	\includegraphics[width=0.99\textwidth]{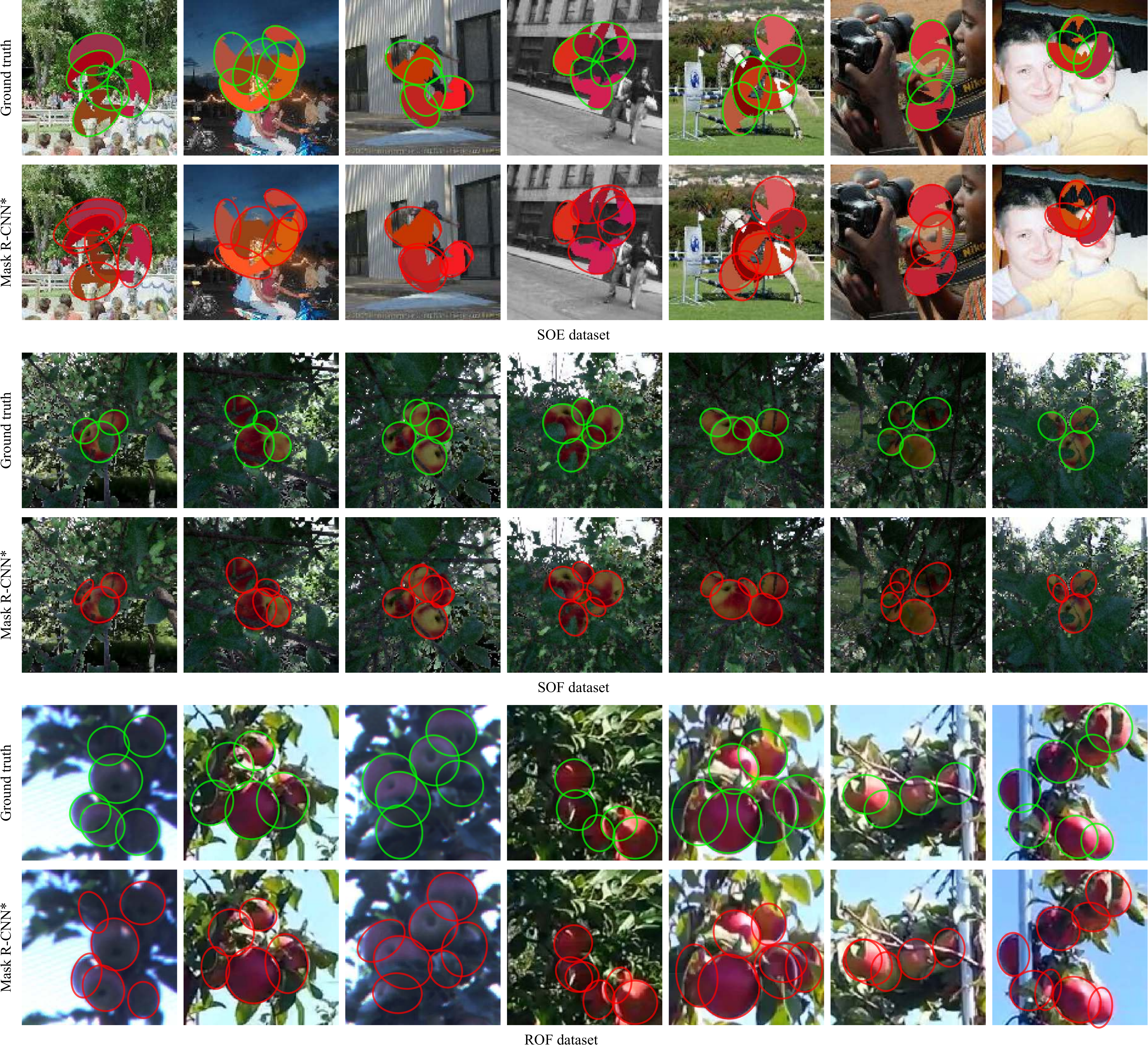}
	\caption{Qualitative results from Mask R-CNN* on SOE, SOF, and ROF datasets. The Mask R-CNN* model trained on the visible parts, to some extent, outputs fewer false positives compared to Mask R-CNN+. However, it cannot correctly capture the entire geometric shapes of objects in heavily occluded scenarios.}
	\label{fig:expMaskVis}
\end{figure*}

In this experiment, we train Mask R-CNN~\cite{hemask} directly on the visible parts of occluded objects (represented as Mask R-CNN*).
Fig.~\ref{fig:expMaskVis} shows the qualitative results from Mask R-CNN*.
We can observe that Mask R-CNN* trained on visible parts, to some extent, outputs fewer false positives compared to Mask R-CNN+.
However, Mask R-CNN* cannot correctly capture the entire shapes of objects in heavily occluded scenarios, which is not suitable for further 3D object localization and size estimation (see Fig.~\ref{fig:motivation3D}).
Compared to Mask R-CNN*, our Ellipse R-CNN is able to accurately detect and infer the entire geometric shapes of occluded objects directly from their visible parts.

\subsection{Extra Evaluation on FDDB Dataset: $\text{AP}_{50:5:95}$}
In Table~\ref{tab:fddbStandard}, we demonstrate the quantitative results from both Mask R-CNN and our Ellipse R-CNN- using the standard AP over pixel IoU ($\text{AP}_{50:5:95}$ without the ellipse angle criterion).
For the FDDB dataset, the ground truth of each face is annotated as an ellipse (instead of a visible region) to represent the entire face shape (see Fig.~\ref{fig:expResFDDB}).
To train Mask R-CNN, we generate the GT mask (in pixel) within the GT ellipse.
For the GT ellipse out of the image, we keep the ellipse pixels that are within the image boundaries, since proposed regions from the region proposal network (RPN) of the Mask R-CNN model cannot be located outside the image.
We train the models using GT ellipses (ellipse parameters for Ellipse R-CNN- and ellipse masks for Mask R-CNN).
We can observe that, in terms of the standard AP over pixel IoU, our Ellipse R-CNN- also achieves better performance than Mask R-CNN.
The reason is that our robust ellipse regression strategy can accurately capture the ellipse parameters (numerically stable) even some faces are occluded by image boundaries such that the generated ellipse masks are also accurate.

\subsection{Performance in terms of MR over Bounding-Box IoU}
To further demonstrate the effectiveness of handling occlusion, we compare our Ellipse R-CNN model with traditional methods of bounding-box detection.
In Table~\ref{tab:bboxIoU}, we show the quantitative results from Mask R-CNN-bbox, DeepParts-bbox~\cite{tian2015deep}, SENet-bbox~\cite{hu2018squeeze}, and our Ellipse R-CNN using the standard MR score over bounding-box IoU ($\text{MR}_{70:5:90}$).
Similar to Faster R-CNN~\cite{ren2017faster}, Mask R-CNN-bbox outputs bounding-box detections.
This baseline model is directly modified from Mask R-CNN by removing the segmentation branch.
Both bounding-box regression and classification are built into Mask R-CNN-bbox, DeepParts-bbox, and SENet-bbox.
We train such three models on the entire object regions (rectangular) for bounding-box detection.
For Ellipse R-CNN, we generate the rectangular enclosing regions of GT ellipses and predicted ellipses to calculate the bounding-box IoU values.
Each model is evaluated on SOE, SOF, and ROF datasets, respectively.
From Table~\ref{tab:bboxIoU}, we can observe that, in terms of MR over bounding-box IoU, all four models  achieve better performances (smaller scores) than those in Table~\ref{tab:ellipse}--\ref{tab:real}.
The reason is that MR over ellipse IoU (ellipse mask) is more strict than MR over bounding-box IoU (no information about ellipse angle) such that the performance score is less affected by rectangular predicted object regions.
Overall, Ellipse R-CNN still achieves the best performance: Mask R-CNN-bbox $<$ DeepParts-bbox $<$ SENet-bbox $<$ Ellipse R-CNN, which further validates that the proposed ellipse regression and learning occlusion strategies enable our Ellipse R-CNN model to correctly capture the entire object regions (bounding boxes or ellipses) and effectively handle various occluded cases.

\begin{table*}[thbp!]
	\caption{Overall performance on SOE, SOF, and ROF datasets using $\text{MR}_{70:5:90}$. $\textbf{R}$: proposed ellipse regression strategy. \newline $\text{MR}_{\star}$: the bounding-box IoU level starts from 0.7 to 0.9 with an interval 0.05.} \label{tab:bboxIoU}
	\begin{center}
		\begin{tabular}{l|M{.22cm} M{.22cm} M{.22cm}|M{.7cm} M{.7cm} M{.7cm}|M{.7cm} M{.7cm} M{.7cm}|M{.7cm} M{.7cm} M{.7cm}}
			\toprule
			\multicolumn{4}{c|}{Datasets} & \multicolumn{3}{c|}{SOE} & \multicolumn{3}{c|}{SOF} & \multicolumn{3}{c}{ROF} \\
			\midrule
			Methods & \textbf{R} & \textbf{O} & \textbf{A} & $\text{MR}_{\star}$ & $\text{MR}_{70}$ & $\text{MR}_{80}$ & $\text{MR}_{\star}$ & $\text{MR}_{70}$ & $\text{MR}_{80}$ & $\text{MR}_{\star}$ & $\text{MR}_{70}$ & $\text{MR}_{80}$ \\
			\bottomrule\toprule
			Mask R-CNN-bbox & -- & -- & -- & 57.9 & 24.2 & 58.6 & 68.8 & 29.7 & 74.3 & 68.5 & 43.7 & 70.3 \\
			DeepParts-bbox~\cite{tian2015deep} &  & \checkmark &  & 40.2 & 13.3 & 29.0 & 64.2 & 22.3 & 68.3 & 62.4 & 28.4 & 58.7 \\
			SENet-bbox~\cite{hu2018squeeze} &  &  & \checkmark & 38.6 & 11.8 & 27.6 & 63.5 & 21.7 & 68.2 & 59.5 & 27.9 & 58.1 \\
			Ellipse R-CNN & \checkmark & \checkmark & \checkmark & \textbf{33.8} & \textbf{6.6} & \textbf{25.1} & \textbf{61.0} & \textbf{17.5} & \textbf{66.8} & \textbf{56.7} & \textbf{23.7} & \textbf{55.0} \\
			\bottomrule
		\end{tabular}
	\end{center}
\end{table*}